\documentclass{article}
\usepackage{graphicx} %

\usepackage[utf8]{inputenc}

\usepackage{cite}

\usepackage{float}

\usepackage{placeins}

\usepackage{appendix}

\usepackage{caption}
\usepackage{subcaption}

\usepackage{makecell}

\usepackage{hyperref}

\usepackage[left=2cm,right=2cm,top=2.5cm,bottom=2.5cm]{geometry}

\usepackage{amsmath}
\usepackage{amssymb}

\newcommand{\eg}{{\em e.g.,~}}
\newcommand{\ie}{{\em i.e.,~}}

\newcommand{\etal}{{\em ~et al.}}

\usepackage{fancyhdr}
\title{Modularity in Deep Learning: A Survey}
\author{Haozhe Sun $^{1}$ \and Isabelle Guyon $^{1,2}$}
\date{$^{1}$ \quad LISN/CNRS/INRIA, Université Paris-Saclay, France\\
$^{2}$ \quad ChaLearn, USA \\
Email: haozhe.sun@universite-paris-saclay.fr}

\begin{document}

\maketitle

\begin{abstract}

Modularity is a general principle present in many fields. It offers attractive advantages, including, among others, ease of conceptualization, interpretability, scalability, module combinability, and module reusability. The deep learning community has long sought to take inspiration from the modularity principle, either implicitly or explicitly. This interest has been increasing over recent years. We review the notion of modularity in deep learning around three axes: data, task, and model, which characterize the life cycle of deep learning. Data modularity refers to the observation or creation of data groups for various purposes. Task modularity refers to the decomposition of tasks into sub-tasks. Model modularity means that the architecture of a neural network system can be decomposed into identifiable modules. We describe different instantiations of the modularity principle, and we contextualize their advantages in different deep learning sub-fields. Finally, we conclude the paper with a discussion of the definition of modularity and directions for future research.

\end{abstract}

\section{Introduction}

Modularity is a general principle present in many fields such as biology~\cite{bongardEvolvingModularGenetic2002, wagnerPerspectiveComplexAdaptations1996, fodorModularityMind1983, robbinsModularityMind2017, barrettModularityCognitionFraming2006, cosmidesCognitiveAdaptationsSocial1992, frankenhuisEvolutionaryPsychologyFodor2007, cosmidesOriginsDomainSpecificity1994, fodorMindDoesnWork2000, pylyshynVisionContinuousCognition1999, kurzweilHowCreateMind2013, pereira-lealOriginsEvolutionFunctional2006, cluneEvolutionaryOriginsModularity2013, hofmanEvolutionHumanBrain2014}, complex systems~\cite{simonArchitectureComplexity1962, simonAggregationVariablesDynamic1961}, mathematics~\cite{avigadModularityMathematics2020, bourbakiArchitectureMathematics1950}, system design~\cite{parnasCriteriaBeUsed1972, gentileTheoryModularityHypothesis2013, shaoModularityMeasuresConcepts2020, modrakDevelopmentModularityMeasure2021, cohen-boulakiaScientificWorkflowsComputational2017}, computer science~\cite{baldwinDesignRulesPower1999, fordArchitectsIntelligenceTruth2018}, graph theory~\cite{newmanModularityCommunityStructure2006, muffLocalModularityMeasure2005, poisotPosterioriMeasureNetwork2013, gomezNewModularityMeasure2016}. While sharing the same name, there is no universally agreed upon definition of modularity~\cite{benaExtremeSparsityGives2021}. However, we can identify a shared definition~\cite{amerReviewModularizationTechniques2019, schmidtModularityConceptNew2001}: in general, modularity is the property of an entity whereby it can be broken down into a number of sub-entities (referred to as modules). This definition has different instantiations in different fields with their nuances~\cite{schillingGeneralModularSystems2000} from which various properties may arise. Such field-specific properties include autonomy of modules (limited interaction or limited interdependence between modules)~\cite{parnasCriteriaBeUsed1972, baldwinDesignRulesPower1999, jacobsAdaptiveMixturesLocal1991, azamBiologicallyInspiredModular2000, newmanModularityCommunityStructure2006, galantiModularityHypernetworks2020, avigadModularityMathematics2020, shaoModularityMeasuresConcepts2020, modrakDevelopmentModularityMeasure2021, cluneEvolutionaryOriginsModularity2013, huizingaEvolvingNeuralNetworks2014, goyalRecurrentIndependentMechanisms2021, yuMAttNetModularAttention2018, aliasparthgoyalNeuralProductionSystems2021}, functional specialization of modules~\cite{fodorModularityMind1983, cosmidesOriginsDomainSpecificity1994, robbinsModularityMind2017, frankenhuisEvolutionaryPsychologyFodor2007, gentileTheoryModularityHypothesis2013, kurzweilHowCreateMind2013, cluneEvolutionaryOriginsModularity2013}, reusability of modules~\cite{avigadModularityMathematics2020, reisingerEvolvingReusableNeural2004, andreasNeuralModuleNetworks2016, aletModularMetalearning2019, csordasAreNeuralNets2021, schmidtModularityConceptNew2001, panDecomposingDeepNeural2020, panDecomposingConvolutionalNeural2021, cohen-boulakiaScientificWorkflowsComputational2017, pontiCombiningModularSkills2022, lakeHumanlevelConceptLearning2015, chenModularMetalearningShrinkage2020, parascandoloLearningIndependentCausal2018, claveraPolicyTransferModularity2017, rahamanDynamicInferenceNeural2021}, combinability of modules~\cite{andreasNeuralModuleNetworks2016, aletModularMetalearning2019, lakeCompositionalGeneralizationMeta2019, parnasCriteriaBeUsed1972, rahamanDynamicInferenceNeural2021, liLearningCompositionalVisual2020a, pontiCombiningModularSkills2022, veniatEfficientContinualLearning2021, mittalCompositionalAttentionDisentangling2022}, replaceability of modules~\cite{parnasCriteriaBeUsed1972, panDecomposingDeepNeural2020, panDecomposingConvolutionalNeural2021}.

As a general principle, modularity is a descriptive property and an organizational scheme. It is a means of representing entities (data, tasks, models) to be able to manipulate them, conceptually or practically~\cite{ghaziRecursiveSketchesModular2019, baldwinDesignRulesPower1999, parnasCriteriaBeUsed1972, cohen-boulakiaScientificWorkflowsComputational2017}. Though modular entities are not necessarily hierarchical~\cite{parnasCriteriaBeUsed1972}, many modular entities have a hierarchical structure~\cite{simonArchitectureComplexity1962} in the sense that multiple modules of a lower hierarchy level can form one module of a higher hierarchy level. The modules of the lower hierarchy level are of finer granularity than those of the higher hierarchy level. At the same level of the hierarchy, modules can refer to an exclusive division of the overall entity (hard division) or overlapping parts of the overall entity (soft division). The decomposed modules can be homogeneous (similar modules) or heterogeneous (dissimilar modules).

\begin{figure}[h]
    \centering
    \includegraphics[width=0.9\linewidth]{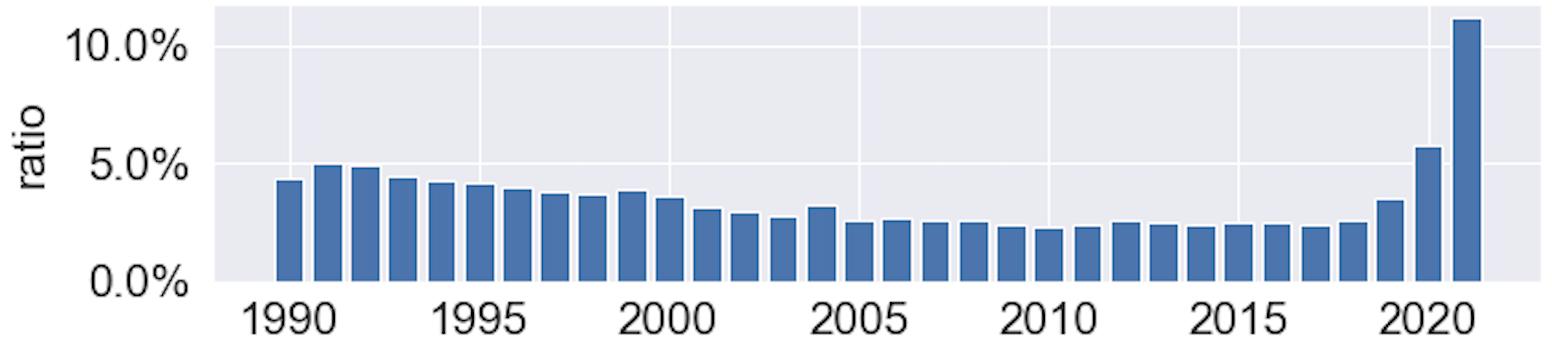}
    \caption{{\bf Publication trend of ``modular deep learning'' from 1990 to 2021.} The ratio of the count of publications containing ``modular deep learning'' and ``modular neural network'' among publications containing ``deep learning'' and ``neural network'', indexed by Google Scholar. The horizontal axis is the publication year.}
    \label{fig:google_scholar_keyword_trends}
    
\end{figure}

\begin{figure}[h]
    \centering
    \includegraphics[width=\linewidth]{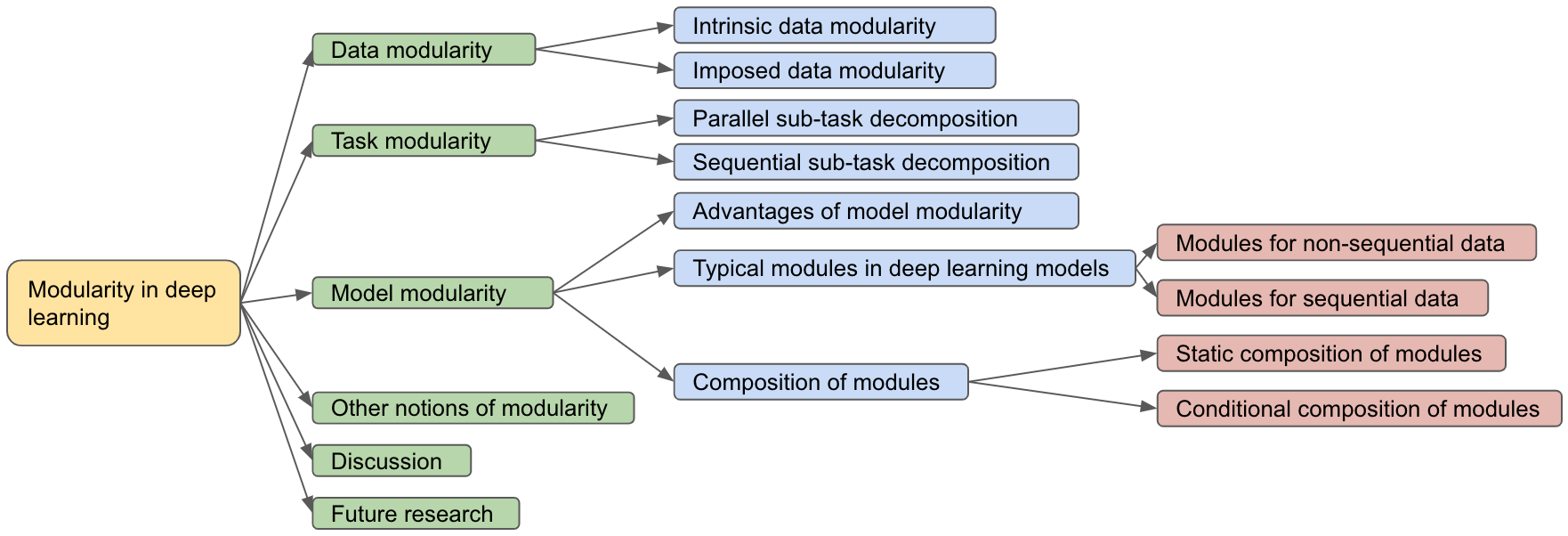}
    \caption{{\bf Organization of this survey.} The first three sections discuss how the modularity principle is instantiated in the three axes: data, task, and model architecture. We then cover other modularity notions for completeness. Finally, we discuss the definition of modularity and directions for future research. The introduction and conclusion are ignored in this figure.}
    \label{fig:sections_plot}
\end{figure}

Back to the very beginning of neural network research in the last century, the community started to be interested in bringing the notion of modularity to neural networks~\cite{audaModularNeuralNetworks1999, azamBiologicallyInspiredModular2000, jacobsAdaptiveMixturesLocal1991, reisingerEvolvingReusableNeural2004}, this interest has been revived recently~\cite{andreasNeuralModuleNetworks2016, shazeerOutrageouslyLargeNeural2017, fernandoPathNetEvolutionChannels2017a, kirschModularNetworksLearning2018, aletModularMetalearning2019, amerReviewModularizationTechniques2019, csordasAreNeuralNets2021, filanClusterabilityNeuralNetworks2021, changModularityReinforcementLearning2021, veniatEfficientContinualLearning2021}. The publication trend (Figure~\ref{fig:google_scholar_keyword_trends}) shows an increasing interest in the modularity principle within deep learning over recent years. This survey investigates the notion of modularity in deep learning around three axes: data, task, and model. The organization of the survey is shown in Figure~\ref{fig:sections_plot}.

\FloatBarrier

\section{Data modularity}

\label{sec:modularityindata}

Data is an entity used to represent knowledge and information. In the context of machine learning and deep learning, it can take various forms \eg image, audio sound, and text. Data samples can be interpreted as points in a high dimensional space (fixed-length dense vectors)~\cite{almeidaWordEmbeddingsSurvey2019, kohComparisonAnalysisDeep2021, krizhevskyImageNetClassificationDeep2012}. A collection of data samples is a dataset. Datasets can be used to train or test deep learning models, referred to as training or test datasets. In these scenarios, data is the input of deep learning models (neural networks)~\cite{goodfellowDeepLearning2016}.

Data modularity is the observation or creation of data groups; it refers to how a dataset can be divided into different modules for various purposes. The division of the dataset into modules facilitates conception and data manipulation. Data modularization can influence the training of learning machines~\cite{hacohenPowerCurriculumLearning2019, elbazLessonsLearnedNeurIPS2022a, sunOmniPrintConfigurablePrinted2021}. Some algorithms leverage data modularity so that each data module is processed by a different solver~\cite{qiaoNovelModularRBF2020}.

We identify two types of data modularity: intrinsic data modularity and imposed data modularity. Intrinsic data modularity means identifiable dataset divisions naturally in data, which a human practitioner does not introduce. Imposed data modularity means identifiable dataset divisions that a human practitioner introduces. The rationale of this taxonomy is that when the dataset reaches the practitioner who analyses it, it already contains some form of intrinsic modularity, including that stemming from the class labels. The people who collect the data are not considered practitioners.

\begin{figure}[h]
    \centering
    \includegraphics[width=0.8\linewidth]{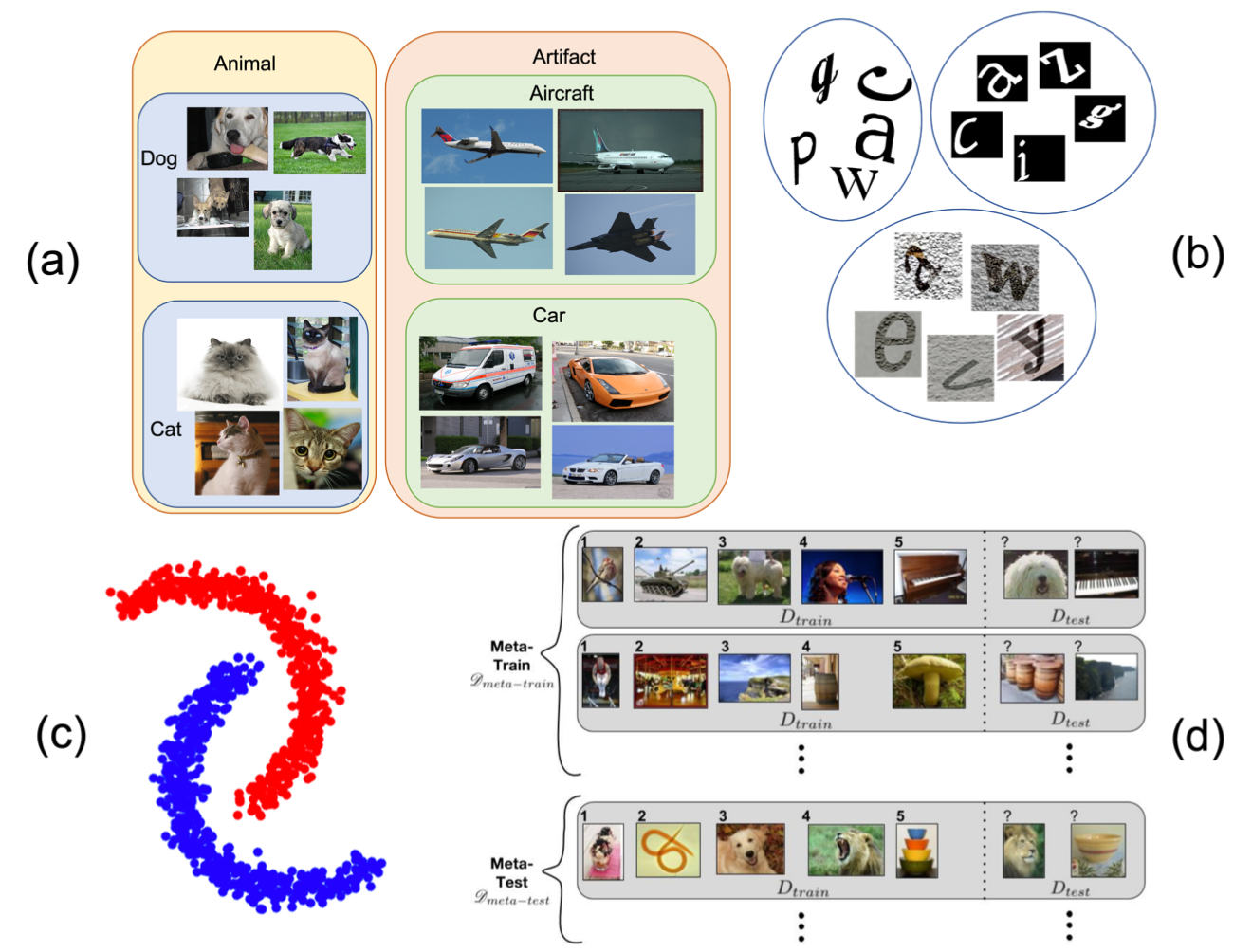}
    \caption{{\bf Illustration of modularity in data.} (a) intrinsic data modularity based on super-classes, images, and class hierarchy in ImageNet~\cite{dengImageNetLargescaleHierarchical2009}; (b) intrinsic data modularity based on styles characterized by a set of metadata, the upper-left circle contains black-on-white characters, the upper-right circle contains white-on-black characters, the lower circle contains characters with natural foreground and background, all characters are drawn from the same set of classes (small-case Latin characters), these three circles illustrate the division of a character dataset based on its metadata; (c) intrinsic manifolds in the form of a moon dataset, where each data manifold can be considered as a module; (d) few-shot learning episodes, reprinted from \cite{raviOptimizationModelFewshot2017}. (a), (b) and (c) are examples of intrinsic data modularity, (d) is an example of imposed data modularity.}
    \label{fig:data_section_summary}

\end{figure}

\subsection{Intrinsic data modularity}

Intrinsic data modularity means identifiable dataset divisions naturally in data, which are not introduced by a human practitioner.

Any supervised learning datasets can be divided according to the classes (labels); data points belonging to the same class are supposed to be close to each other in a hidden space, which allows for solutions of classification algorithms. Classes sharing common semantics can be further grouped to form super-classes. For example, ImageNet~\cite{dengImageNetLargescaleHierarchical2009} has a class hierarchy (see Figure~\ref{fig:data_section_summary} (a)) which is used by Meta-Dataset~\cite{triantafillouMetadatasetDatasetDatasets2019}. Omniglot dataset~\cite{lakeHumanlevelConceptLearning2015} and OmniPrint datasets~\cite{sunOmniPrintConfigurablePrinted2021} contain character images organized in scripts, each script (super-class) contains several characters (classes); Meta-Album dataset~\cite{ullahMetaAlbumMultidomainMetaDataset2022} is a meta-dataset including 40 datasets, where each dataset can be considered as a super-class. The super-classes provide information about class similarity, allowing splitting datasets according to the semantics~\cite{yosinskiHowTransferableAre2014}.

In addition to the classes or super-classes, data points can also be grouped by one or several metadata such as time, location, and gender. Such metadata is available with the Exif data of photos. The OmniPrint data synthesizer generates data together with a comprehensive set of metadata, including font, background, foreground, margin size, shear angle, rotation angle, etc.~\cite{sunOmniPrintConfigurablePrinted2021} (see Figure~\ref{fig:data_section_summary} (b)). The NORB dataset collected stereo image pairs of 50 uniform-colored toys under 36 angles, 9 azimuths, and 6 lighting conditions, where the angles, azimuths, and lighting conditions serve as the metadata~\cite{lecunLearningMethodsGeneric2004}.

Some datasets contain intrinsic clusters in the high-dimensional feature space. Such intrinsic clusters can stem from the underlying data generative process, where latent categorical variables determine the natural groups of data. An illustrative example is a Gaussian Mixture distribution where data points are assumed to be generated from a mixture of a finite number of Gaussian distributions with unknown parameters~\cite{hastieElementsStatisticalLearning2009}. Some datasets have intrinsic manifolds; an illustrative example is the moons dataset as shown in Figure~\ref{fig:data_section_summary} (c), where the two manifolds interlace while preserving an identifiable division, each manifold can be considered as a module. Both of the above examples fall into the category of data clustering. When data samples are interconnected in the form of a graph~\cite{maDisentangledGraphConvolutional2019, wuLearningImplicitSemantic2021}, this is called graph partitioning. One question which arises is how to determine the optimal clustering of a dataset. Luxburg\etal~\cite{luxburgClusteringScienceArt2012} argue that there are no optimal domain-independent clustering algorithms and that clustering should always be studied in the context of its end-use.

Multi-modal deep learning aims to build models that can process and relate information from multiple modalities. Here the modality refers to the way in which something happens or is experienced \eg data in the form of image, text, audio~\cite{baltrusaitisMultimodalMachineLearning2018}. Multi-modal datasets fall into the category of intrinsic data modularity in the sense that the data in each modality can be considered a module. For example, VQA v2.0 dataset~\cite{goyalMakingVqaMatter2017} consists of open-ended questions about images; SpeakingFaces dataset~\cite{abdrakhmanovaSpeakingfacesLargescaleMultimodal2021} consists of aligned thermal and visual spectra image streams of fully-framed faces synchronized with audio recordings of each subject speaking.

\subsection{Imposed data modularity}

Imposed data modularity means identifiable dataset divisions which are introduced by a human practitioner. 

When training deep learning models~\cite{goodfellowDeepLearning2016}, human practitioners usually divide the whole training dataset into mini-batches, which can be seen as a kind of imposed data modularity. The gradient is computed using one mini-batch of data for each parameter update; one training epoch means passing through all the mini-batches. This iterative learning regime is called stochastic gradient descent~\cite{ruderOverviewGradientDescent2016}. Mini-batches reduce the memory requirement for backpropagation, which makes training large deep learning models possible. On the other hand, batch size also influences learning behavior. Smith\etal~\cite{smithDonDecayLearning2018} showed that the benefits of decaying the learning rate could be obtained by instead increasing the training batch size. Keskar\etal~\cite{keskarLargeBatchTrainingDeep2017} showed that learning with large batch sizes usually gives worse generalization performance.

Instead of using a sequence of mini-batches sampled uniformly at random from the entire training dataset, curriculum learning~\cite{hacohenPowerCurriculumLearning2019} uses non-uniform sampling of mini-batches such that the mini-batch sequence exhibits an increasing level of difficulty. A related concept is active learning~\cite{renSurveyDeepActive2021}, which assumes that different data points in a dataset have different values for the current model update; it tries to select the data points with the highest value to construct the actual training set.

The model performance is usually tested on few-shot episodes in few-shot learning and meta-learning. Few-shot episodes are typically formed by drawing several classes \(N\) from the class pool and several examples \(K\) for each selected class, called \(N\)-way-\(K\)-shot episodes~\cite{finnModelAgnosticMetaLearningFast2017, snellPrototypicalNetworksFewshot2017} (Figure~\ref{fig:data_section_summary} (d)). For such scenarios, the meta-training phase can employ the same episodic learning regime or not~\cite{triantafillouMetadatasetDatasetDatasets2019}, recent studies~\cite{wangBridgingMultitaskLearning2021, wangRoleGlobalLabels2021, laenenEpisodesPrototypicalNetworks2021} and competition results~\cite{elbazLessonsLearnedNeurIPS2022a} suggest that episodic meta-training is not more effective than vanilla pretraining with access to the global class pool.

Data augmentation is a way to generate more training data by applying transformations to existing data~\cite{simardBestPracticesConvolutional2003}. The transformed versions of the same data point can be seen as a module. Some transformations, such as rotation and translation, form a group structure~\cite{roseCourseGroupTheory1994}. The effect of such data augmentation can be understood as averaging over the orbits of the group that keeps the data distribution approximately invariant and leads to variance reduction~\cite{chenGrouptheoreticFrameworkData2020}.

In addition to splitting the dataset into subsets of samples, each data sample can be split into subdivisions of features, referred to as feature partitioning. A dataset can be represented as a matrix where each row represents one data sample; each column represents one feature dimension. It can then be divided along the sample and feature dimensions. Schmidt\etal~\cite{schmidtModularityConceptNew2001} process each feature partition with a different model. For image classification tasks, input images can be split into small patches that can be processed in parallel~\cite{jinSplitCNNSplittingWindowBased2019, dosovitskiyImageWorth16x162021}.

\subsection{Conclusion of data modularity}

We argue that data without structure contains no useful information for learning dependencies (\eg between feature and label). Some dependencies boil down to the emergence or the creation of groups. Intrinsic data modularity relates to the semantic relationship between samples and how data samples are similar or dissimilar. Imposed data modularity, on the other hand, relates to the way that practitioners organize data at hand to better train learning machines. 

Future research for data-centric deep learning may investigate the relationship between intrinsic and imposed data modularity. For example, does intrinsic data modularity promote imposed data modularity? How does this interplay affect model training?

Data modularity describes how the input of deep learning models can be modularized. On the other hand, the end goal (the output) of deep learning models can also be modularized, which is the topic of the next section.

\FloatBarrier

\section{Task modularity}

\label{sec:modularityintask}

\begin{figure}[h]
    \centering
    \includegraphics[width=0.5\linewidth]{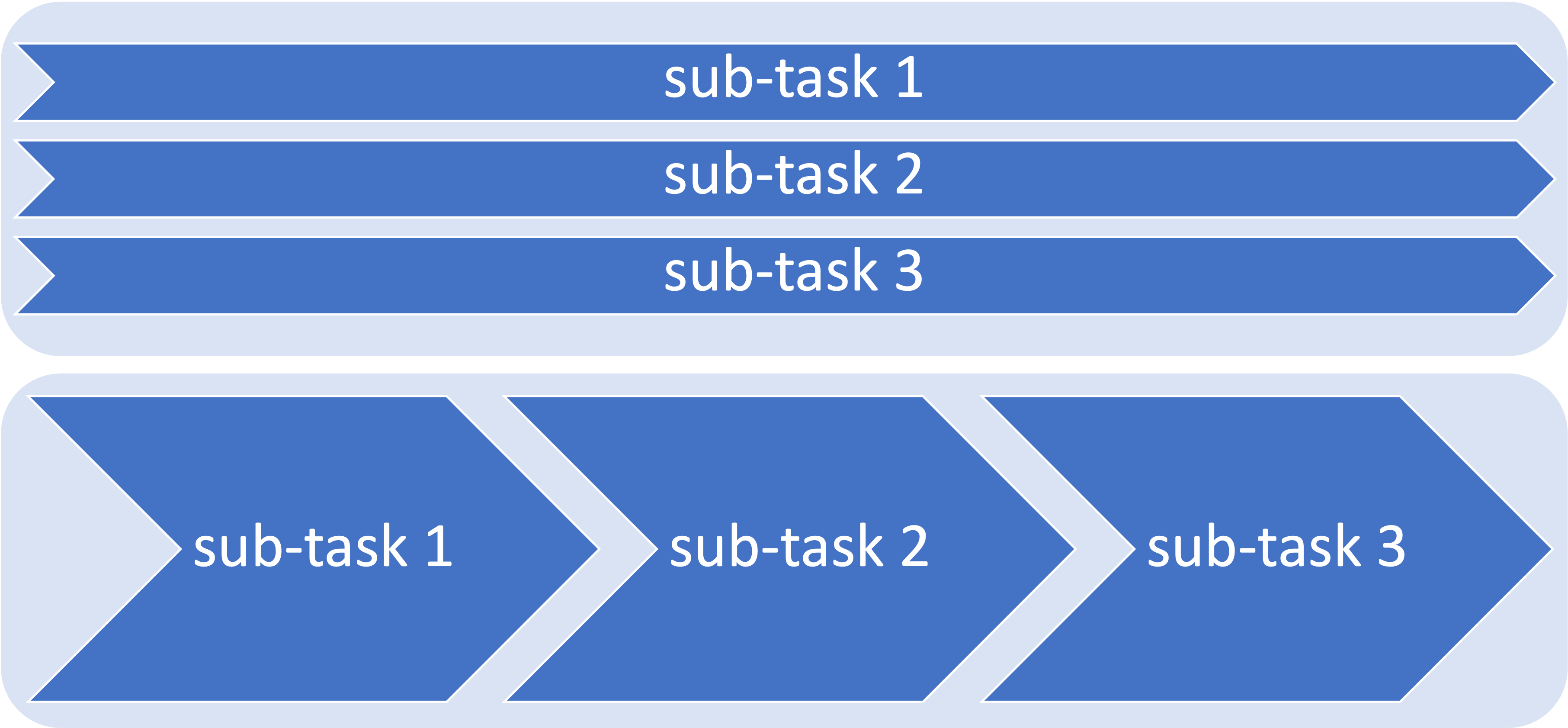}
    \caption{{\bf Illustration of sub-task decomposition.} The upper figure illustrates the parallel decomposition of a task. The lower figure illustrates the sequential decomposition of a task.}
    \label{fig:sub_task_decomposition}

\end{figure}

Deep learning models are tools to solve tasks \eg from the classification of entities to the generation of realistic photos. Solving a task is equal to achieving a corresponding objective. In deep learning, we usually model an objective by an explicit differentiable objective function (also known as a loss function), allowing end-to-end training. This perspective can be generalized to any task, even if the objective function is implicit and does not entail a differentiable form. For example, the task of ``purchasing a cup of tea'' can be characterized by an indicator function that returns a penalty if no tea can be purchased or a bonus otherwise. In deep learning, tasks are often related to data; but they are different. Given the same dataset, one can define various tasks on top of it. For example, the MNIST dataset can be used either for an image classification benchmarking task~\cite{mazziaEfficientcapsnetCapsuleNetwork2021} or for a pixel sequence classification benchmarking task~\cite{kruegerZoneoutRegularizingRNNs2017, goyalRecurrentIndependentMechanisms2021}, the OmniPrint-meta[1-5] datasets~\cite{sunOmniPrintConfigurablePrinted2021} can be used either for a few-shot learning benchmarking task or for domain adaptation benchmarking task. Tasks define the objective; they are orthogonal to how the end goal should be achieved.

This section presents task modularity \ie sub-task decomposition. Sub-task decomposition means that a task could be factorized or decomposed into sub-tasks. Sub-task decomposition facilitates conceptualization and problem-solving. The divide-and-conquer principle breaks down a complex problem into easier sub-problems~\cite{cormenIntroductionAlgorithms2022, qiaoNovelModularRBF2020, azamBiologicallyInspiredModular2000, jacobsTaskDecompositionCompetition1991}. By solving each individual sub-problem and combining the solutions, the complex problem can be solved more efficiently. The sub-task decomposition facilitates the integration of expert knowledge, and the a priori knowledge can further facilitate problem-solving. Sub-task decomposition can also promote reuse if the overall task is compositional; the solution to sub-tasks may be reused in other tasks~\cite{ostapenkoAttentionCompositionalModularity2022, meyersonModularUniversalReparameterization2019, devinLearningModularNeural2017, simpkinsComposableModularReinforcement2019, purushwalkamTaskdrivenModularNetworks2019}.

The sub-task decomposition can be categorized into two regimes: parallel decomposition and sequential decomposition (Figure~\ref{fig:sub_task_decomposition}). Parallel decomposition means that the sub-tasks can be executed in parallel. Sequential decomposition means that the sub-tasks need to be executed in order; certain sub-tasks cannot be executed before the previous sub-task is finished. In practice, these two regimes can be mixed. For example, a sub-task from a sequential decomposition can be further decomposed parallelly, which leads to a directed acyclic graph workflow.

\subsection{Parallel sub-task decomposition}

A parallel sub-task decomposition is called homogeneous if the decomposed sub-tasks are similar. One typical example is dividing a multi-class classification problem into multiple smaller classification problems~\cite{ghaziRecursiveSketchesModular2019}. Given a neural network trained to perform a multi-class classification problem, Csordás\etal~\cite{csordasAreNeuralNets2021} use parameter masks to identify subsets of parameters solely responsible for individual classes on their own. Kim\etal~\cite{kimSplitNetLearningSemantically2017} learn to split a neural network into a tree structure to handle different subsets of classes. They assume that different classes use different features, the tree-structured neural network ensuring that the later layers do not share features across different subsets of classes. Pan\etal~\cite{panDecomposingDeepNeural2020, panDecomposingConvolutionalNeural2021} and Kingetsu\etal~\cite{kingetsuNeuralNetworkModule2021} decompose a multi-class classification model into reusable, replaceable and combinable modules, where each module is a binary classifier. Such modules can be recombined without retraining to obtain a new multi-class classifier. These methods can be useful in situations where the classes to be classified frequently change. Abbas\etal~\cite{abbasDeTraCTransferLearning2020} use transfer learning and class decomposition to improve the performance of medical image classification. Such sub-task decomposability is an implicit prerequisite of the model editing problem~\cite{sinitsinEditableNeuralNetworks2019, mitchellFastModelEditing2021, kassnerBeliefBankAddingMemory2021, mitchellMemoryBasedModelEditing2022, mengLocatingEditingFactual2022}. Model editing aims to modify a specific sub-task learned by a trained neural network without damaging model performance on other inputs, \eg it aims to patch the mistake of the model for a particular sample. If the task cannot be decomposed into disentangled sub-tasks, then model editing cannot be achieved.

A parallel sub-task decomposition is termed heterogeneous if the decomposed sub-tasks are dissimilar; such decomposition is usually problem-dependent and requires expert knowledge of the task at hand. Belay\etal~\cite{belayFactoredConvolutionalNeural2019} decompose the recognition task of Amharic characters into a vowel recognition task and a consonant recognition task to reduce overall task complexity. Cao\etal~\cite{caoDeFormerDecomposingPretrained2020} decompose the full self-attention into question-wide and passage-wide self-attentions to speed up inference for question answering tasks. Ding\etal~\cite{dingTrunkbranchEnsembleConvolutional2017} decompose the facial recognition task into multiple facial component recognition tasks. Zhou\etal~\cite{zhouMetaLearningSymmetriesReparameterization2020} decompose the neural network learning task into structure learning and parameter learning to learn equivariance from data automatically. Gatys\etal~\cite{gatysImageStyleTransfer2016} decompose the natural image synthesis task into a content component and a style component, which allows recombining the content and the style in a combinatorial way to generate new images.

\subsection{Sequential sub-task decomposition}

Sequential sub-task decomposition reflects the sequential pipeline of the task. A simple example is the division of a machine learning task into a preprocessing stage (data cleaning and normalization) and a model inference stage~\cite{ranganathanStudyFindFacts2021}.

In reinforcement learning, a complex task can usually be decomposed~\cite{simpkinsComposableModularReinforcement2019} into a sequence of sub-tasks or steps. An illustrative example is to imagine that the task of manufacturing an artifact \(Z\) requires purchasing the raw material \(X\), forging \(X\) to produce parts \(Y\), and then assembling the parts \(Y\) into the end product \(Z\). Both \(X\) and \(Y\) can take different values independently (\(X \in \{x_1, x_2, x_3, ...\}, Y \in \{y_1, y_2, y_3, ...\}\)). Different values of \(X\) and \(Y\) can be recombined, which forms a combinatorial number of possible scenarios to learn. This pipeline can be factorized into three stages: (1) raw material purchase, (2) forging to produce parts, and (3) assembling of parts. Reinforcement learning agents would learn more efficiently if the learning happens at the granularity of the factorized stages instead of the overall task~\cite{colasCuriousIntrinsicallyMotivated2019}. Furthermore, such a factorization enables the independence of credit assignment~\cite{petersElementsCausalInference2017}; the failure of the overall task can be traced back to the problematic stages, while the other stages can remain untouched. For example, if the raw material is of bad quality, then the purchase sub-task needs to be improved; the forging sub-task and the assembling sub-task do not need to be changed~\cite{changModularityReinforcementLearning2021}.

The sequential pipeline is omnipresent in practical applications \eg optical character recognition (OCR), natural language processing (NLP). When facing a multi-script (multi-language) recognition task, the pipeline can consist of a script identification stage and a script-specific recognition stage~\cite{shiScriptIdentificationWild2016, huangMultiplexedNetworkEndtoend2021}, which decouples the domain classifier and the domain-specific solver. The text-in-the-wild recognition task~\cite{chenTextRecognitionWild2020} usually consists of decoupled text detector (to localize the bounding box of the text) and recognizer (recognize the text from the bounding box)~\cite{chenTextRecognitionWild2020}. Traditional OCR methods also decompose the word recognition task into a character segmentation task and a character recognition task~\cite{caseySurveyMethodsStrategies1996, schenkelRecognitionbasedSegmentationOnline1992, choudharyNewCharacterSegmentation2013, kaurStudyVariousCharacter2015}. Traditional NLP pipeline includes sentence segmentation, word tokenization, part-of-speech tagging, lemmatization, filtering stop words, and dependency parsing~\cite{jurafskySpeechLanguageProcessing2019}. In bioinformatics, the scientific workflow (data manipulations and transformations) groups similar or strongly coupled workflow steps into modules to facilitate understanding and reuse~\cite{cohen-boulakiaScientificWorkflowsComputational2017}.

\subsection{Conclusion of task modularity}

The sub-task decomposition can be parallel, sequential, or mixed (directed acyclic graph). We provided examples from the literature that leverage sub-task decomposition to reduce task complexity or promote the reuse of sub-task solutions. Task modularity can help integrate expert knowledge and promote model interpretability when paired with model modularity, as will be discussed in the next section.

Future research may focus on how to automate the process of sub-task decomposition or make the problem-dependent sub-task decomposition techniques transferable to other tasks, which is an important step for AutoML. It would reduce the demand for highly qualified deep learning engineers, which can reduce expert bias and entry barriers to deep learning.

\FloatBarrier

\section{Model modularity}

\label{sec:modularityinmodel}

This section presents model modularity. It means that the architecture of the neural network system (one neural network or a system of neural networks) consists of identifiable sub-entities (modules). 

Model modularity is different from task modularity. A task define an objective, task modularity focuses on decomposing the objective into sub-objectives. Model modularity focuses on the architecture of the neural network system, it decomposes the solution into sub-solutions.

\subsection{Advantages of model modularity}
\label{sec:advantagesofmodularityinmodel}

Model modularity provides ease of conceptual design and implementation. For example, modern neural networks consist of repeated layer/block patterns (modules). Examples include fully-connected neural networks~\cite{goodfellowDeepLearning2016}, vanilla convolutional neural networks, ResNet~\cite{heDeepResidualLearning2016, xieAggregatedResidualTransformations2017}, Inception~\cite{szegedyRethinkingInceptionArchitecture2016} and models searched by Neural Architecture Search (NAS)~\cite{zophLearningTransferableArchitectures2018, elskenNeuralArchitectureSearch}. The design with homogeneous modules allows for a more concise description of the model architecture in the sense of Kolmogorov complexity (short description length)~\cite{liIntroductionKolmogorovComplexity2008, lecunOptimalBrainDamage1989}. For example, instead of specifying how each primitive operation (\eg sum, product, concatenation) interacts in a computational graph, the model can be described as a collection of modules that interact with each other~\cite{ghaziRecursiveSketchesModular2019}. The standardization of such neural network building blocks (fully-connected layers, convolutional layers) also enabled the development of highly optimized hardware and software ecosystems for fast computation~\cite{paszkePyTorchImperativeStyle2019, martinabadiTensorFlowLargeScaleMachine2015, AccelerateFastMath, gavaliChapterDeepConvolutional2019, grayGPUKernelsBlockSparse}.

Together with sub-task decomposition (task modularity), model modularity offers ease of expert knowledge integration~\cite{andreasNeuralModuleNetworks2016, goodfellowGenerativeAdversarialNetworks2014, silverMasteringGameGo2016, shinContinualLearningDeep2017, belayFactoredConvolutionalNeural2019} and interpretability~\cite{kirschModularNetworksLearning2018, pontiCombiningModularSkills2022, jacobsTaskDecompositionCompetition1991, krishnamurthyInterpretabilityGatedModular2021}. Interpretability can have different forms. For example, each neural network module could be assigned a specific interpretable sub-task. On the other hand, selective module evaluation provides insights on how different samples/tasks are related~\cite{jacobsAdaptiveMixturesLocal1991, shazeerOutrageouslyLargeNeural2017, andreasNeuralModuleNetworks2016, aletModularMetalearning2019} in the context of conditional computation~\cite{bengioEstimatingPropagatingGradients2013}.

The model decomposition into modules promotes reusability and knowledge transfer~\cite{braylanReuseNeuralModules2016}. Though each neural network is typically trained to perform a specific task, its (decomposed) modules could be shared across tasks if appropriate mechanisms promote such reusability. The simplest example would be the classical fine-tuning paradigm of large pretrained models~\cite{hintonReducingDimensionalityData2006, yosinskiHowTransferableAre2014, chuBestPracticesFinetuning2016, heRethinkingImagenetPretraining2019}. This paradigm typically freezes the pretrained model and only retrains its last classification layer to adapt it to the downstream task. Pretrained models are typically pretrained on large datasets~\cite{russakovskyImageNetLargeScale2015, sunRevisitingUnreasonableEffectiveness2017, yalnizBillionscaleSemisupervisedLearning2019a}. The large amount and diversity of training data make pretrained models' intermediate features reusable for other downstream tasks. More recently, the finer-grained reusability of neural network systems has attracted the attention of researchers. Such methods assume that the tasks share underlying patterns and keep an inventory of reusable modules (each module is a small neural network)~\cite{andreasNeuralModuleNetworks2016, kirschModularNetworksLearning2018, aletModularMetalearning2019, veniatEfficientContinualLearning2021}. Each module learns different facets (latent factors or atomic skills) of the knowledge required to solve each task. The selective/sparse use and dynamic reassembling/recombination of these modules can promote sample efficiency~\cite{pontiCombiningModularSkills2022} and combinatorial generalization~\cite{andreasNeuralModuleNetworks2016, damarioHowModularShould2021, aletModularMetalearning2019, islamDiscreteFactorialRepresentations2022}.

Combinatorial generalization is also known as compositional generalization, ``infinite use of finite means''~\cite{chomskyAspectsTheorySyntax1965}, and systematic generalization. It aims to generalize to unseen compositions of known functions/factors/words~\cite{csordasCTLEvaluatingGeneralization2022, fodorConnectionismCognitiveArchitecture1988, lakeGeneralizationSystematicityCompositional2018, keysersMeasuringCompositionalGeneralization2020, ostapenkoAttentionCompositionalModularity2022, changAutomaticallyComposingRepresentation2019, purushwalkamTaskdrivenModularNetworks2019}, it is the ability to systematically recombine previously learned elements to map new inputs made up from these elements to their correct output~\cite{schmidhuberCompositionalLearningDynamic1990}. For example, new sentences consist of new compositions of a known set of words. Combinatorial generalization is argued to be important to achieve human-like generalization~\cite{battagliaRelationalInductiveBiases2018, pontiCombiningModularSkills2022, mittalModularArchitectureEnough2022, lecunPathAutonomousMachine2022, hupkesCompositionalityDecomposedHow2020, pontiInductiveBiasModular2021, kingetsuNeuralNetworkModule2021, wangCombinatorialPerspectiveTransfer2020, lakeHumanlevelConceptLearning2015, hupkesStateoftheartGeneralisationResearch2022, vankovTrainingNeuralNetworks2020, loulaRearrangingFamiliarTesting2018}. Learning different facets of knowledge with different modules in a reusable way could be one solution to combinatorial generalization. Modular systems have been shown effective for combinatorial generalization~\cite{rosenbaumRoutingNetworksChallenges2019} in various fields \eg natural language processing~\cite{lakeCompositionalGeneralizationMeta2019, pontiCombiningModularSkills2022, hupkesCompositionalityDecomposedHow2020, pontiInductiveBiasModular2021, murtyCharacterizingIntrinsicCompositionality2022}, visual question answering~\cite{andreasNeuralModuleNetworks2016, damarioHowModularShould2021, bahdanauSystematicGeneralizationWhat2019}, object recognition~\cite{purushwalkamTaskdrivenModularNetworks2019, lakeHumanlevelConceptLearning2015, parascandoloLearningIndependentCausal2018, changAutomaticallyComposingRepresentation2019}, and robotics~\cite{aletModularMetalearning2019, devinLearningModularNeural2017, pathakLearningControlSelfassembling2019, claveraPolicyTransferModularity2017}.

The modularization of neural network systems promotes knowledge retention. If different knowledge is localized into different modules, targeted knowledge updates and troubleshooting~\cite{kingetsuNeuralNetworkModule2021, panDecomposingDeepNeural2020, panDecomposingConvolutionalNeural2021} will be possible. This can alleviate gradient interference of different tasks~\cite{yuGradientSurgeryMultiTask2020, kanakisReparameterizingConvolutionsIncremental2020, maninisAttentiveSingleTaskingMultiple2019} and catastrophic forgetting~\cite{veniatEfficientContinualLearning2021, rusuProgressiveNeuralNetworks2016, terekhovKnowledgeTransferDeep2015, andersonFineTuningModular2016, javedMetaLearningRepresentationsContinual2019, frenchUsingSemiDistributedRepresentations1991, aletModularMetalearning2019, abrahamMemoryRetentionSynaptic2005, ellefsenNeuralModularityHelps2015, keAchievingForgettingPrevention2021, ostapenkoContinualLearningLocal2021}.

Modular neural network systems facilitate model scaling in two ways. (1) Modular models like fully-connected models and ResNet can be scaled up (or down) by simply stacking more (or less) modules to increase (or decrease) the model capacity to fit larger (or smaller) datasets~\cite{heDeepResidualLearning2016}. (2) Modular methods based on sparsely activated Mixture-of-Experts~\cite{shazeerOutrageouslyLargeNeural2017} decouple computation cost from model size. They allow drastically increasing the model capacity without increasing compute cost because only a small fraction of the model is evaluated on each forward pass~\cite{fedusReviewSparseExpert2022, shazeerOutrageouslyLargeNeural2017, heFasterMoEModelingOptimizing2022, duGlamEfficientScaling2022, chowdheryPaLMScalingLanguage2022, barhamPathwaysAsynchronousDistributed2022}. The extreme example of these sparsely activated models is Switch Transformer~\cite{fedusSwitchTransformersScaling2022} which contains 1.6 trillion parameters, pushing the competition of large model sizes~\cite{brownLanguageModelsAre2020, smithUsingDeepSpeedMegatron2022} to the next level.

\subsection{Typical modules in deep learning models}

\label{sec:typicalmodules}

This section reviews some typical modules in the deep learning literature. 

Almost all systems are modular to some degree~\cite{schillingGeneralModularSystems2000}, neural network systems can almost always be decomposed into subsystems (modules)~\cite{balestrieroPOLICEProvablyOptimal2022} following different points of view. More specifically, they usually consist of a hierarchical structure in which a module of a higher hierarchy level is made of modules of a lower hierarchy level. The elementary layer of modern neural networks (\eg fully-connected layer, convolutional layer) can be seen as a module on its own. On the other hand, any neural network as a whole can also be considered as a module \eg in the context of ensemble~\cite{zhouEnsembleMethodsFoundations2012}, Mixture-of-Experts~\cite{jacobsAdaptiveMixturesLocal1991}, and Generative Adversarial Networks (GAN)~\cite{goodfellowGenerativeAdversarialNetworks2014}. Some literature~\cite{csordasAreNeuralNets2021, benaExtremeSparsityGives2021, sunTaskSwitchingNetwork2021, kingetsuNeuralNetworkModule2021} define modules as sub-neural networks where part of the parameters are masked out (set to 0). In these cases, overlapping modules can be obtained when the masks overlap.

\begin{figure}[h]
    \centering
    \includegraphics[width=\linewidth]{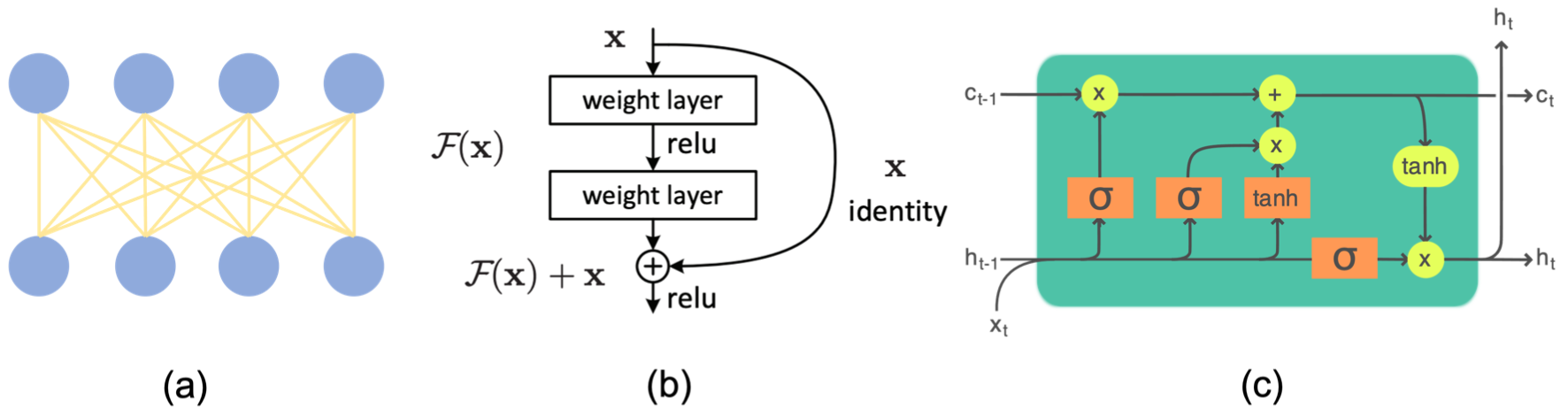}
    \caption{{\bf Examples of a module.} (a) a fully-connected layer; (b) a basic ResNet module, reprinted from \cite{heDeepResidualLearning2016}; (c) an LSTM module, reprinted from \cite{chevalierLongShorttermMemory2022}.}
    \label{fig:module_examples}

\end{figure}

\noindent
{\bf 4.2.1 Modules for non-sequential data}

\noindent
Fully-connected layers (Figure~\ref{fig:module_examples} (a)) imitate the connections between neurons in biological neural networks but connect every input neuron to every output neuron~\cite{goodfellowDeepLearning2016}. In practice, a fully-connected layer is implemented as a matrix multiplication between input data and learnable parameters. Convolutional layers introduce the inductive bias of translation equivariance. Conceptually, a convolutional layer (with a single output channel) can be obtained from a fully-connected layer by enforcing local connectivity and parameter sharing~\cite{goodfellowDeepLearning2016}. Local connectivity means that each neuron only connects to a subset of neurons of the previous layer; parameter sharing means that the same learnable parameters are used across receptive fields. In practice, a convolutional layer is implemented as a collection of kernels/filters shifted over the input data~\cite{paszkePyTorchImperativeStyle2019, martinabadiTensorFlowLargeScaleMachine2015}. Each kernel performs a dot product between input data and learnable parameters. Depending on the number of dimensions over which kernels are shifted, a convolutional layer is termed \eg 1D, 2D, 3D. 2D convolutional layers are widely used in computer vision tasks~\cite{lecunGradientbasedLearningApplied1998, krizhevskyImageNetClassificationDeep2012}. Locally connected layers are similar to convolutional layers except that they remove the constraint of parameter sharing (across kernels). It helps if one wants to impose local receptive fields while there is no reason to think each local kernel should be the same~\cite{goodfellowDeepLearning2016}. Low-rank locally connected layers relax spatial equivariance and provide a trade-off between locally connected layers and convolutional layers. The kernel applied at each position is constructed as a linear combination of a basis set of kernels with spatially varying combining weights. Varying the number of basis kernels allows controlling the degree of relaxation of spatial equivariance~\cite{elsayedRevisitingSpatialInvariance2020}. Standard convolutional layers offer translation equivariance; a line of research focuses on generalizing this to other equivariances (rotation, reflection), referred to as group convolutional layers~\cite{cohenGroupEquivariantConvolutional2016, dielemanExploitingCyclicSymmetry2016, cohenSteerableCNNs2016, worrallHarmonicNetworksDeep2017, gaoEfficientInvariantConvolutional2017, bekkersRotoTranslationCovariantConvolutional2018, weilerLearningSteerableFilters2018, weilerGeneralEquivariantSteerable2021}. On the other hand, depthwise separable convolutional layers~\cite{sifreRigidmotionScatteringImage2014, cholletXceptionDeepLearning2017, howardMobilenetsEfficientConvolutional2017} factorize a standard convolutional layer into a depthwise convolutional layer and a pointwise convolutional layer, which reduces model size and computation.

Multiple layers can be grouped into a building block (a module of a higher hierarchy level). Such examples include the building blocks of ResNet~\cite{heDeepResidualLearning2016}, Inception~\cite{szegedyRethinkingInceptionArchitecture2016, szegedyInceptionv4InceptionresnetImpact2017}, ResNeXt~\cite{xieAggregatedResidualTransformations2017}, Wide ResNet~\cite{zagoruykoWideResidualNetworks2016}. Inception~\cite{szegedyRethinkingInceptionArchitecture2016, szegedyInceptionv4InceptionresnetImpact2017} has parallel kernels of multiple sizes within each block and merge their results to extract information at varying scales. Inception also includes several techniques to reduce computation cost \eg factorizing large kernels into smaller kernels and using \(1\times 1\) convolution to reduce dimensionality. A ResNet block~\cite{heDeepResidualLearning2016} (Figure~\ref{fig:module_examples} (b)) contains a sequence of convolutional layers; it adds a skip-connection (also known as residual connection, identity mapping) from the beginning to the end of the block to alleviate vanishing gradients. Many variants of the ResNet block have been proposed. For example, Wide ResNet~\cite{zagoruykoWideResidualNetworks2016} increases the block width; ResNeXt~\cite{xieAggregatedResidualTransformations2017} aggregates parallel paths within each block.

The block design could be automatically searched instead of handcrafted. In order to narrow down the model search space, some Neural Architecture Search methods~\cite{elskenNeuralArchitectureSearch, hutterAutomaticMachineLearning2019, zophLearningTransferableArchitectures2018, yingNasbench101ReproducibleNeural2019} automatically search the optimal design pattern for a block (also known as a cell) while fixing the block composition scheme (also known as meta-architecture). Once the block design patterns are searched, the full model is instantiated by repeating the searched blocks following the predefined block composition scheme. For example, NAS-Bench-101~\cite{yingNasbench101ReproducibleNeural2019} defines the block search space as all possible directed acyclic graphs on V nodes (\(V\leqslant 7\)) while limiting the maximum number of edges to 9.

McNeely-White\etal~\cite{mcneely-whiteInceptionResNetFeatures2020} report that the features learned by Inception and ResNet are almost linear transformations of each other, even though these two architectures have a remarkable difference in the architectural design philosophy. This result explains why the two architectures usually perform similarly and highlights the importance of training data. This result is corroborated by Bouchacourt\etal~\cite{bouchacourtGroundingInductiveBiases2021}, who argue that invariance generally stems from the data itself rather than from architectural bias.

\noindent
{\bf 4.2.2 Modules for sequential data}

\noindent
When the input data is sequential \eg time series, text, audio, video, Recurrent Neural Networks (RNN)~\cite{rumelhartLearningInternalRepresentations1985} come into play. The RNN module processes the sequential data one at a time; the output (also known as the hidden state) of the RNN module at the previous time step is recursively fed back to the RNN module, which allows it to aggregate information across different time steps. The vanilla RNN module suffers from short-term memory issues; it cannot effectively preserve information over long sequences. To overcome this issue, gated recurrent unit (GRU)~\cite{choPropertiesNeuralMachine2014} and long short-term memory (LSTM)~\cite{hochreiterLongShorttermMemory1997} module use gates to control which information should be stored or forgotten in the memory, which allows better preservation of long-term information. In GRU and LSTM modules, gates are neural networks with trainable parameters. While GRU modules are faster to train than LSTM modules, their performance comparison varies depending on the scenario. GRU surpasses LSTM in long text and small dataset scenarios while LSTM outperforms GRU in other scenarios~\cite{yangLSTMGRUNeural2020}.

Contrary to RNN, GRU, and LSTM, which process sequential data one at a time, self-attention layers~\cite{vaswaniAttentionAllYou2017} process the data sequence in parallel. For each data point in a data sequence (\eg each time step of a time series), a self-attention layer creates three transformed versions, referred to as query vector, key vector, and value vector, through linear transformations. Between each pair of data points, the dot product between the query vector and the key vector of the pair reflects how much those two data points are related within the sequence. These dot products are then normalized and combined with the corresponding value vectors to get the new representation of each data point in the sequence. An enhanced version of self-attention layers is multi-head self-attention layers, which extract different versions of query vector, key vector, and value vector for each data point. Multi-head self-attention layers improve performance by capturing more diverse representations. A transformer block combines multi-head self-attention layers, fully-connected layers, normalization layers, and skip-connections. Models built upon transformer blocks have achieved state-of-the-art performance in a wide range of tasks such as natural language processing~\cite{kentonBERTPretrainingDeep2019} and speech synthesis~\cite{liNeuralSpeechSynthesis2019}. Transformer models can be applied to image modality by transforming each input image into a sequence of small image patches~\cite{dosovitskiyImageWorth16x162021}. Despite the lack of image-specific inductive bias (translation equivariance, locality), vision transformers can achieve state-of-the-art performance when combined with a large amount of training data~\cite{dosovitskiyImageWorth16x162021, heMaskedAutoencodersAre2022, baoBEiTBERTPreTraining2022}.

\FloatBarrier

\subsection{Composition of modules}

\label{sec:compositionofmodules}

\begin{figure}[h]
    \centering
    \includegraphics[width=0.6\linewidth]{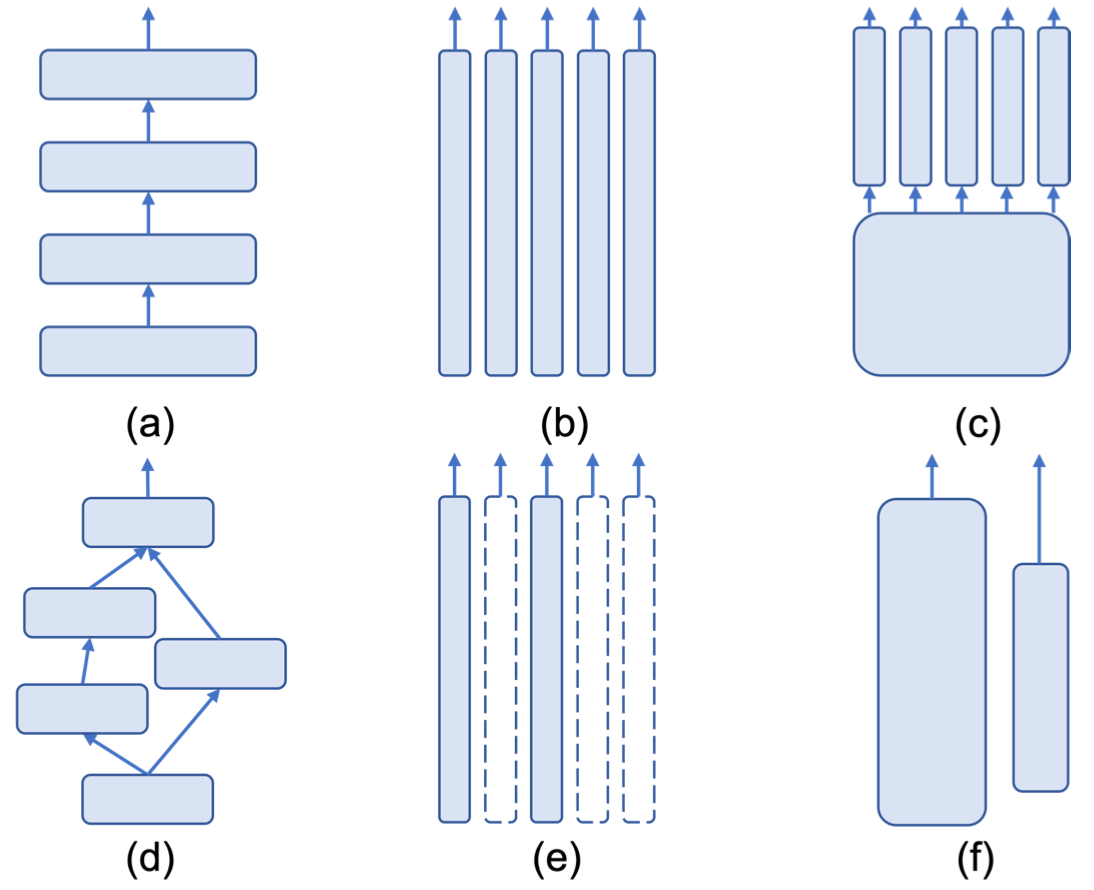}
    \caption{{\bf Illustration of module composition.} (a) Sequential concatenation. (b) Ensembling. (c) Tree-structure composition. (d) General Directed Acyclic Graph. (e) Conditional composition. (f) Cooperation composition.}
    \label{fig:module_composition}

\end{figure}

Section~\ref{sec:typicalmodules} presents typical modules in the literature. Section~\ref{sec:compositionofmodules} discusses how to organize these modules to form a model (or a module of a higher hierarchy level).

\noindent
{\bf 4.3.1 Static composition of modules}

\noindent
Static composition means that the composed structure does not vary with input; the same structure is used for all input samples or tasks. 

One straightforward way to compose modules is sequential concatenation (Figure~\ref{fig:module_composition} (a)). It implies that multiple (typically homogeneous) modules are sequentially concatenated into a chain to form a model, where a module's output is the next module's input. Examples of sequential concatenation include fully-connected models~\cite{goodfellowDeepLearning2016} and ResNet models~\cite{heDeepResidualLearning2016}. This composition scheme typically does not assume an explicit sub-task decomposition; the chain of concatenated modules can instead be seen as a series of information extraction steps~\cite{yosinskiHowTransferableAre2014, tishbyDeepLearningInformation2015, alainUnderstandingIntermediateLayers2016}, extracted features transition from low-level to high-level.

Ensembling composition~\cite{zhouEnsembleMethodsFoundations2012, juRelativePerformanceEnsemble2018, opitzEfficientModelAveraging2016}, on the other hand, organizes modules in a parallel manner (Figure~\ref{fig:module_composition} (b)). The principle of ensembling is to aggregate (\eg averaging) the results of multiple modules (weaker learners) to obtain a more robust prediction. The rationale is that different modules are expected to provide complementary and diverse views of input data. Each module's data is processed independently without relying on the other modules at inference time. The regularization method Dropout~\cite{srivastavaDropoutSimpleWay2014}, which randomly deactivates neurons during training, can be seen as an implicit ensemble method of overlapping modules.

Sequential composition and parallel composition can be combined, \eg in the form of a tree structure (Figure~\ref{fig:module_composition} (c)). A typical scenario of tree-structure composition is a model with a shared feature extractor and multiple task-specific heads~\cite{zhangSurveyMultitaskLearning2021, silverMasteringGameGo2017}. All the above composition schemes are special cases of DAG (Directed Acyclic Graph, Figure~\ref{fig:module_composition} (d)). The general DAG composition scheme is typically found in models searched by Neural Architecture Search~\cite{liuHierarchicalRepresentationsEfficient2018, xieExploringRandomlyWired2019, reisingerEvolvingReusableNeural2004}.

Cooperation composition (Figure~\ref{fig:module_composition} (f)) assumes that each module is a standalone neural network with specific functionality and that these neural networks cooperate during training or inference; it is a neural network system that consists of multiple separate neural networks. Different from ensembling composition, modules in cooperation composition are typically heterogeneous and interact with each other more diversely. For example, siamese networks~\cite{bromleySignatureVerificationUsing1994, chenExploringSimpleSiamese2021, jingMaskedSiameseConvNets2022} consists of two neural networks (module) which work together to produce different versions of the input data. Generative Adversarial Networks (GAN)~\cite{goodfellowGenerativeAdversarialNetworks2014, zhuUnpairedImagetoImageTranslation2017} trains a generator under the guidance of a discriminator. The same spirit applies to teacher and student neural networks~\cite{tarvainenMeanTeachersAre2018}. Some deep reinforcement learning methods implement the Actor-Critic~\cite{suttonReinforcementLearningIntroduction2018} with two separate new networks, such as AlphaGo~\cite{silverMasteringGameGo2016}, A3C~\cite{mnihAsynchronousMethodsDeep2016}, ACKTR~\cite{wuScalableTrustregionMethod2017}. Continual learning with deep replay buffer~\cite{shinContinualLearningDeep2017} consists of a continual neural network learner and a generative neural network serving as the replay buffer. Some other continual learning methods~\cite{rusuProgressiveNeuralNetworks2016, terekhovKnowledgeTransferDeep2015, andersonFineTuningModular2016, veniatEfficientContinualLearning2021} continuously expanding model capacity for new tasks by adding new modules which work in cooperation with old modules.

\noindent
{\bf 4.3.2 Conditional composition of modules}

\noindent
Conditional composition (Figure~\ref{fig:module_composition} (e)) is complementary to static composition in the sense that the composed modules are selectively (conditionally, sparsely, or dynamically) activated (used or evaluated) for each particular input. The input conditioning can happen at the granularity of individual sample~\cite{andreasNeuralModuleNetworks2016, kirschModularNetworksLearning2018, jacobsAdaptiveMixturesLocal1991} as well as task~\cite{pontiCombiningModularSkills2022, maninisAttentiveSingleTaskingMultiple2019, sunTaskSwitchingNetwork2021, masseAlleviatingCatastrophicForgetting2018, jacobsTaskDecompositionCompetition1991}. In the literature, this paradigm is also termed conditional computation~\cite{bengioEstimatingPropagatingGradients2013, bengioConditionalComputationNeural2016}.

The idea of conditional computation can be traced back to Mixture-of-Experts (MoE) introduced in the last century. An MoE is a system composed of multiple separate neural networks (modules), each of which learns to handle a sub-task of the overall task~\cite{jacobsTaskDecompositionCompetition1991, zhouDiverseEnsembleEvolution2018} \eg a subset of the complete training dataset. A gating network computes the probability of assigning each example to each module~\cite{jacobsAdaptiveMixturesLocal1991, jordanHierarchicalMixturesExperts1994} or a sparse weighted combination of modules~\cite{shazeerOutrageouslyLargeNeural2017}. Two issues of MoE are module collapse~\cite{shazeerOutrageouslyLargeNeural2017, kirschModularNetworksLearning2018, mittalModularArchitectureEnough2022} and shrinking batch size~\cite{shazeerOutrageouslyLargeNeural2017}, both of which are related to the balance of module utilization. Module collapse means under-utilization of modules or lack of module diversity. Due to the self-reinforcing behavior of the gating network during training, premature modules may be selected and thus trained even more. The gating network may end up converging to always selecting a small subset of modules while the other modules are never used. Shrinking batch size means the batch size is reduced for each conditionally activated module. Large batch sizes are necessary for modern hardware to make efficient inferences because they alleviate the cost of data transfers~\cite{shazeerOutrageouslyLargeNeural2017}.

\begin{figure}[h]
    \centering
    \includegraphics[width=0.7\linewidth]{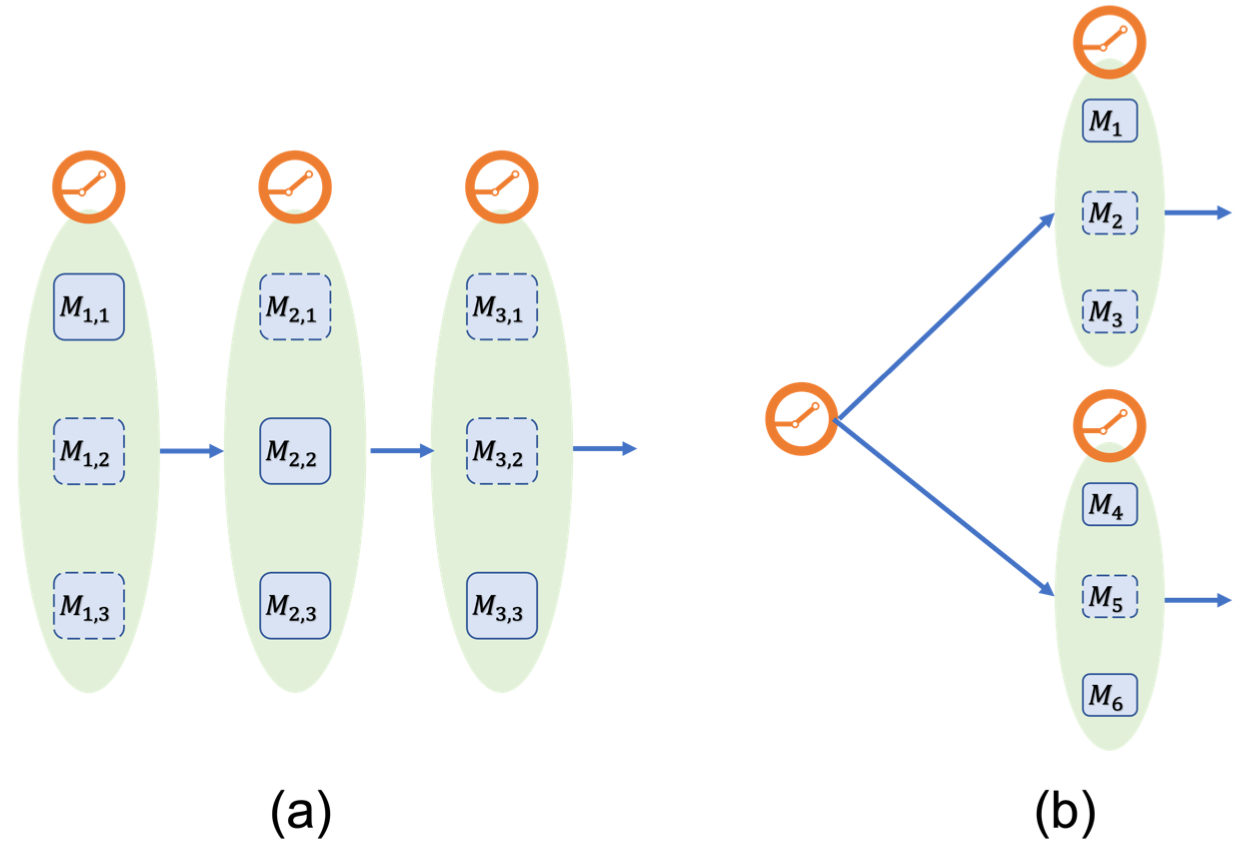}
    \caption{{\bf Extension of Mixture-of-Experts (MoE).} (a) A stacked MoE, which stacks multiple MoE layers into a chain. (b) A hierarchical MoE, where a primary gating network chooses a sparse weighted combination of ``modules'', each of which is an MoE with its own gating network.}
    \label{fig:MoE_stacked_hierarchical}

\end{figure}

MoE can be generalized to \eg stacked MoE~\cite{eigenLearningFactoredRepresentations2014, kirschModularNetworksLearning2018, rosenbaumRoutingNetworksAdaptive2018, fernandoPathNetEvolutionChannels2017a, ramachandranDiversityDepthPerexample2019} or hierarchical MoE~\cite{shazeerOutrageouslyLargeNeural2017, yaoHierarchicalMixtureClassification2009} (Figure~\ref{fig:MoE_stacked_hierarchical}). Eigen\etal~\cite{eigenLearningFactoredRepresentations2014} first explored stacked MoE; they introduced the idea of using multiple MoE with their own gating networks. In order to train stacked MoE, Kirsch\etal~\cite{kirschModularNetworksLearning2018} use generalized Viterbi Expectation-Maximization algorithm, Rosenbaum\etal~\cite{rosenbaumRoutingNetworksAdaptive2018} employ a multi-agent reinforcement learning algorithm, Fernando\etal~\cite{fernandoPathNetEvolutionChannels2017a} use a genetic algorithm. MoE systems do not always have explicit gating networks; for instance, Fernando\etal~\cite{fernandoPathNetEvolutionChannels2017a} rely on the results of the genetic algorithm to decide the module routing scheme.

Inspired by MoE, some deep learning methods keep an inventory of reusable specialized modules that can be conditionally reassembled for each input. This approach has been advocated to promote knowledge transfer, sample efficiency, and generalization. For example, in visual question answering, Neural Module Networks~\cite{andreasNeuralModuleNetworks2016, huLearningReasonEndtoEnd2017, damarioHowModularShould2021} dynamically reassemble modules into a neural network to locate the attention (region of interest) on the questioned image. The question's parsing guides the reassembling process so that the reassembled model reflects the structure and semantics of the question. For this particular task, the compositionality of modules comes from the compositionality of visual attention. Following the question's syntax, the reassembled modules sequentially modify the attention onto the questioned image. For example, the module associated with the word ``cat'' locates the image region containing a cat, and the module associated with the word ``above'' shifts up the attention. Zhang\etal~\cite{zhangNetworkTransplanting2018} investigated adding new abilities to a generic network by directly transplanting the module corresponding to the new ability, dubbed network transplanting.

Some work relies on the hypothesis that the tasks at hand share some commonalities \ie hidden factors are shared across tasks. Each hidden factor can be learned by a separate module from the module inventory for transfer learning and meta-learning. For example, Alet\etal~\cite{aletModularMetalearning2019} use simulated annealing to meta-learn an inventory of modules reusable across tasks to achieve combinatorial generalization. The parameters of an inventory of modules are optimized during meta-training; the trained modules are reassembled during the meta-test with an optional parameter fine-tuning process. They demonstrated the utility of their method for robotics tasks. Ponti\etal~\cite{pontiCombiningModularSkills2022} assume that each task is associated with a subset of latent discrete skills from a skill inventory. They try to generalize more systematically to new tasks by disentangling and recombining different facets of knowledge. More precisely, they jointly learn a skill-specific parameter vector for each latent skill and a binary task-skill allocation matrix. For each new task, the new model's parameter vector is created as the average of the skill-specific parameter vectors corresponding to the skills present in the new task (in addition to a shared base parameter vector).

The conditional composition scheme also has other forms. For example, Teerapittayanon\etal~\cite{teerapittayanonBranchyNetFastInference2016} save computation on easy input data via early exiting; later layers will be skipped if the intermediate feature's prediction confidence passes a predefined threshold. Fuengfusin\etal~\cite{fuengfusinNetworkSubnetworksLayerwise2020} train models whose layers can be removed at inference time without significantly reducing the performance to allow adaptive accuracy-latency trade-off. Similarly, Yu\etal~\cite{yuSlimmableNeuralNetworks2019} train models which are executable at customizable widths (the number of channels in a convolutional layer). Xiong\etal~\cite{xiongConditionalConvolutionalNeural2015} sparsely activate convolutional kernels within each layer for each particular input sample, which provides an example of the conditional composition of overlapping modules.

\subsection{Conclusion of model modularity}

Section~\ref{sec:modularityinmodel} presents how the notion of modularity is instantiated in the architecture of neural network systems. The structure of neural network modules (Section~\ref{sec:typicalmodules}) and the way to organize the modules (Section~\ref{sec:compositionofmodules}) provide a complementary view of model modularity.

While all modern neural networks are modular to some extent, different instantiations of the modularity principle offer different advantages (Section~\ref{sec:advantagesofmodularityinmodel}). The advantages include ease of conceptual design and implementation, ease of expert knowledge integration, better interpretability, ease of knowledge transfer and reuse, better generalization and sample efficiency, ease of knowledge retention, ease of troubleshooting, and better scalability.

\FloatBarrier

\section{Other notions of modularity}

There remain some other notions of modularity in the deep learning literature.

In graph theory, the term ``modularity'' refers to a measure commonly used in community detection. It measures the density of connections within a community (module) compared to between modules communities~\cite{newmanModularityCommunityStructure2006}. This measure can be applied to graph clustering problems in the form of modularity optimization~\cite{brandesModularityClustering2007, huDeepStockRepresentation2018, salha-galvanModularityAwareGraphAutoencoders2022, shiokawaFastAlgorithmModularitybased2013}. Inspired by this measure, Filan\etal~\cite{filanClusterabilityNeuralNetworks2021} investigate the parameter clustering pattern that emerged from the training of a neural network. They view a neural network as an undirected weighted graph (edge weights are the absolute value of network parameters) and apply spectral clustering on the obtained graph. They observe that some neural networks trained on image classification tasks have some clustering properties of their parameters: edge weights are stronger within one cluster than between clusters. Watanabe\etal~\cite{watanabeModularRepresentationLayered2018} have obtained similar results. Béna\etal~\cite{benaExtremeSparsityGives2021} adapted the graph-theoretic modularity measure to define structural modularity and define functional specialization through three heuristic measures. The functional specialization can be intuitively understood as the extent to which a sub-network can do a sub-task independently. To investigate the relationship between structural and functional modularity, they design a scenario where a model with two parallel modules (with an adjustable number of interconnections) is used to predict whether the parity of the two digits is the same or different. They show that enforcing structural modularity via sparse connectivity between two communicating modules does lead to functional specialization of the modules. However, this phenomenon only happens at extreme levels of sparsity. With even a moderate number of interconnections, the modules become functionally entangled. Mittal\etal~\cite{mittalModularArchitectureEnough2022} observed that modular systems (weighted combination of parallel modules) with a good module specialization are good in terms of the overall system performance, however end-to-end training itself is not enough to achieve a good module specialization.

\begin{figure}[h]
    \centering
    \includegraphics[width=0.6\linewidth]{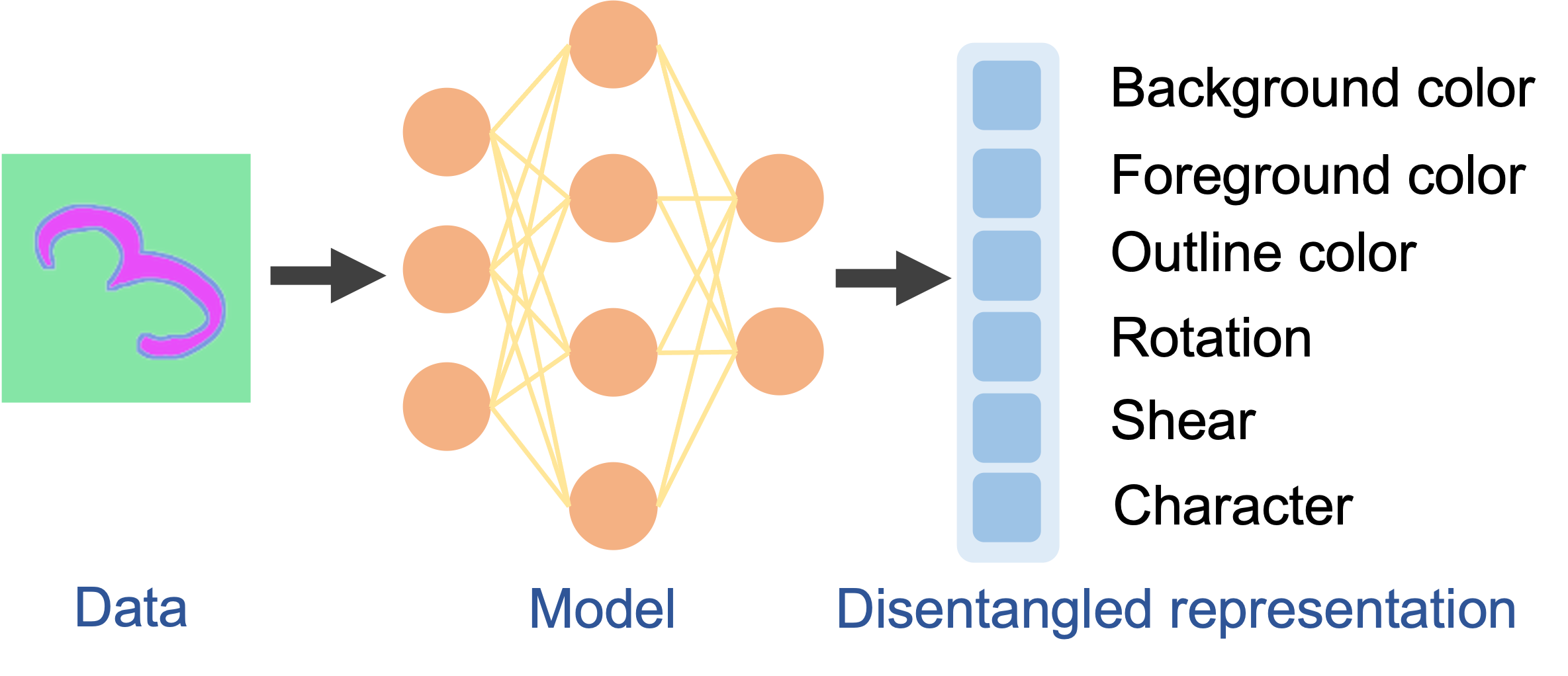}
    \caption{{\bf Illustration of a disentangled representation.} }
    \label{fig:disentangled_representation}

\end{figure}

The term ``modularity'' is related to the notion of independence in some literature. For example, Galanti\etal~\cite{galantiModularityHypernetworks2020} use modularity to refer to the ability of hypernetworks~\cite{haHyperNetworks2016} to learn a different function for each input instance. A line of research has been carried out on learning disentangled representation. Intuitively, disentangled representation aims to reverse the underlying data generating process and retrieve its latent factors into the learned representation (Figure~\ref{fig:disentangled_representation}). One of the desirable properties of a disentangled representation~\cite{ridgewayLearningDeepDisentangled2018, eastwoodFrameworkQuantitativeEvaluation2018, zaidiMeasuringDisentanglementReview2021} is ``modularity''. In this context, a modular representation is a representation where each dimension of the representation conveys information about at most one latent generative factor.

\FloatBarrier

\section{Discussion}

Defining modularity is, in itself, a challenging problem. The notion of modularity is present in literature across many different fields~\cite{bongardEvolvingModularGenetic2002, wagnerPerspectiveComplexAdaptations1996, fodorModularityMind1983, robbinsModularityMind2017, barrettModularityCognitionFraming2006, cosmidesCognitiveAdaptationsSocial1992, frankenhuisEvolutionaryPsychologyFodor2007, cosmidesOriginsDomainSpecificity1994, fodorMindDoesnWork2000, pylyshynVisionContinuousCognition1999, kurzweilHowCreateMind2013, pereira-lealOriginsEvolutionFunctional2006, simonArchitectureComplexity1962, simonAggregationVariablesDynamic1961, avigadModularityMathematics2020, bourbakiArchitectureMathematics1950, parnasCriteriaBeUsed1972, gentileTheoryModularityHypothesis2013, shaoModularityMeasuresConcepts2020, modrakDevelopmentModularityMeasure2021, cohen-boulakiaScientificWorkflowsComputational2017, baldwinDesignRulesPower1999, fordArchitectsIntelligenceTruth2018, newmanModularityCommunityStructure2006, muffLocalModularityMeasure2005, poisotPosterioriMeasureNetwork2013, gomezNewModularityMeasure2016}. While many researchers have a strong intuition about what it means for an entity to be modular, there has yet to be a universal agreement on what defines modularity. The same is true even within the field of deep learning. As rightly said by Béna\etal~\cite{benaExtremeSparsityGives2021}: ``Modularity of neural networks is a bit like the notion of beauty in art: everyone agrees that it's important, but nobody can say exactly what it means''. We argue that the difficulty of defining modularity stems from the fact that the notion of modularity usually comes with many different properties: replaceability of modules, combinability of modules, reusability of modules, autonomy of modules (limited interaction or limited interdependence between modules), functional specialization of modules. Authors from different fields typically only retain one or two of the above properties to claim an entity to be modular.

In this survey, we define modularity as the property of an entity whereby it can be broken down into a number of sub-entities (referred to as modules). This definition is the prerequisite of the properties mentioned above; it is the greatest common definition of the notion of modularity. By recursively applying the definition of modularity, a modular entity is an entity that can be broken down into sub-entities, where each sub-entity can be further broken down into sub-sub-entities. This recursion can be repeated for discrete entities until the atomic elements (minimum indivisible modules) are reached. In that case, a set of atomic elements \(\{a \in D\}\) can formally characterize a discrete entity \(D\); a subset of atomic elements can then characterize a module \(M \subseteq D\). The above framework applies to data modularity (Section~\ref{sec:modularityindata}) and model modularity (Section~\ref{sec:modularityinmodel}). The reason is that data and models are both discrete: data samples and model parameters are stored in physical computers where everything is represented quantitatively. On the other hand, we need to use a different framework for task modularity because tasks are usually not discrete. As discussed in Section~\ref{sec:modularityintask}, each task can be characterized by an objective function \(F\). In this sense, task modularity can be formally characterized by (objective) function compositions. A task is decomposable if there exists a set of functions \(\{f_1, f_2, ...\}\) that, when composed together, retrieve the form of the original objective function \(F\).

For discrete entities, one needs to choose the atomic elements. Naively, one could choose each data sample in a dataset and each neuron in a neural network as the atomic elements. However, both choices remain to be discussed because they are indeed not the smallest indivisible modules. Regarding data modularity, the dataset division can happen both at the sample dimension and the feature dimension, which means that each data sample can be divided into smaller elements \eg feature vectors of reduced length or image patches. Regarding model modularity, the modularization can happen at the granularity of parameters \eg modules can be obtained by masking out parameters~\cite{csordasAreNeuralNets2021, benaExtremeSparsityGives2021, sunTaskSwitchingNetwork2021, kingetsuNeuralNetworkModule2021}. Consequently, one can choose the scalar numbers stored in physical computers (often represented by floating-point numbers) as the atomic elements. The atomic elements for data are every single dimension of data samples; the atomic elements for models are every single scalar parameters in the neural network. It entails that, in some cases, there needs to be some relationship \(R\) among atomic elements \(\{a \in D\}\) because any arbitrary subsets of atomic elements do not necessarily form a valid module if the relationship \(R\) is broken. In the above example, the relationship \(R\) indicates which scalar numbers should come together to form data samples or how to use each scalar parameter along the feedforward computation in the computational graph of neural networks. In consequence, an entity can be a set or a system; a system is a set equipped with relationships \(R\) among atomic elements \(\{a \in D\}\).

\section{Future research}

As modularity is a general principle, this survey covered many elements from different sub-fields of deep learning; each sub-field can provide a lot of future avenues of research on its own. To name a few, McNeely-White\etal~\cite{mcneely-whiteInceptionResNetFeatures2020} and Bouchacourt\etal~\cite{bouchacourtGroundingInductiveBiases2021} showed that given the same training data, learned features exhibit similar properties across models with markedly different architectural inductive biases. Is it still worth improving neural network architectures if data dominate learning results? Future research may validate the results of McNeely-White\etal~\cite{mcneely-whiteInceptionResNetFeatures2020} and Bouchacourt\etal~\cite{bouchacourtGroundingInductiveBiases2021} by extending their research to more kinds of models and training datasets in a more systematic way. If these results still hold, one may need to ground these results theoretically. On the other hand, whether neural networks can learn and behave compositionally is still an open question~\cite{hupkesCompositionalityDecomposedHow2020, andreasMeasuringCompositionalityRepresentation2019}. It entails that we need a domain-agnostic way to test the compositionality of neural networks.

Different aspects of the modularity principle can be further investigated to improve deep learning models. It boils down to designing new deep learning methods that provide \eg better interpretability, reusability, scalability, and efficiency. While model modularity may, to some extent, reflect task modularity, it is still unclear whether data modularity directly corresponds with model modularity. One research avenue is to automate imposed data modularization regarding specific models in the spirit of AutoML. Similarly, automating task modularization can facilitate problem-solving and reduce human-introduced bias.

\section{Conclusion}

Deep learning is becoming dominant in many applications, such as computer vision and natural language processing. It is natural to ask ourselves whether there are guidelines for designing deep learning algorithms. Modularity is one guiding principle that has been put forward in the literature. This survey reveals that modularity is pervasive in three related yet distinct axes of deep learning: data, task, and model architecture. We observed that some modularity concepts come in the form of a prior, while others come in the form of a posterior. 

The efforts of bringing the modularity principle into deep learning are not new; however, reviewing deep learning literature using the point of view of modularity is relatively new. This survey provides a step towards clarifying and investigating the notion of modularity in deep learning and elsewhere.

\section*{Acknowledgments}

We gratefully acknowledge constructive feedback and suggestions from Birhanu Hailu Belay, Romain Egele, Felix Mohr, Hedi Tabia, and the reviewers. This work was supported by ChaLearn and the ANR (Agence Nationale de la Recherche, National Agency for Research) under AI chair of excellence HUMANIA, grant number ANR-19-CHIA-0022.

{
\small
\bibliography{refs}

\begin{thebibliography}{100}

\bibitem{AccelerateFastMath}
Accelerate {{Fast Math}} with {{Intel}}\textregistered{} {{oneAPI Math Kernel
  Library}}.
\newblock
  https://www.intel.com/content/www/us/en/developer/tools/oneapi/onemkl.html.

\bibitem{abbasDeTraCTransferLearning2020}
Asmaa Abbas, Mohammed Abdelsamea, and Mohamed Gaber.
\newblock {{DeTraC}}: {{Transfer Learning}} of {{Class Decomposed Medical
  Images}} in {{Convolutional Neural Networks}}.
\newblock {\em IEEE Access}, PP:1--1, April 2020.

\bibitem{abdrakhmanovaSpeakingfacesLargescaleMultimodal2021}
Madina Abdrakhmanova, Askat Kuzdeuov, Sheikh Jarju, Yerbolat Khassanov, Michael
  Lewis, and Huseyin~Atakan Varol.
\newblock Speakingfaces: {{A}} large-scale multimodal dataset of voice commands
  with visual and thermal video streams.
\newblock {\em Sensors}, 21(10):3465, 2021.

\bibitem{abrahamMemoryRetentionSynaptic2005}
Wickliffe~C. Abraham and Anthony Robins.
\newblock Memory retention \textendash{} the synaptic stability versus
  plasticity dilemma.
\newblock {\em Trends in Neurosciences}, 28(2):73--78, February 2005.

\bibitem{alainUnderstandingIntermediateLayers2016}
Guillaume Alain and Yoshua Bengio.
\newblock Understanding intermediate layers using linear classifier probes.
\newblock {\em arXiv preprint arXiv:1610.01644}, 2016.

\bibitem{aletModularMetalearning2019}
Ferran Alet, Tom{\'a}s {Lozano-P{\'e}rez}, and Leslie~P. Kaelbling.
\newblock Modular meta-learning.
\newblock {\em arXiv:1806.10166 [cs, stat]}, May 2019.

\bibitem{aliasparthgoyalNeuralProductionSystems2021}
Anirudh~Goyal ALIAS PARTH~GOYAL, Aniket Didolkar, Nan~Rosemary Ke, Charles
  Blundell, Philippe Beaudoin, Nicolas Heess, Michael~C Mozer, and Yoshua
  Bengio.
\newblock Neural production systems.
\newblock In M.~Ranzato, A.~Beygelzimer, Y.~Dauphin, P.S. Liang, and J.~Wortman
  Vaughan, editors, {\em Advances in Neural Information Processing Systems},
  volume~34, pages 25673--25687. {Curran Associates, Inc.}, 2021.

\bibitem{almeidaWordEmbeddingsSurvey2019}
Felipe Almeida and Geraldo Xex{\'e}o.
\newblock Word {{Embeddings}}: {{A Survey}}, January 2019.

\bibitem{amerReviewModularizationTechniques2019}
Mohammed Amer and Tomas Maul.
\newblock A {{Review}} of {{Modularization Techniques}} in {{Artificial Neural
  Networks}}.
\newblock {\em Artificial Intelligence Review}, 52, June 2019.

\bibitem{andersonFineTuningModular2016}
Ark Anderson, Kyle Shaffer, Artem Yankov, Court~D. Corley, and Nathan~O. Hodas.
\newblock Beyond {{Fine Tuning}}: {{A Modular Approach}} to {{Learning}} on
  {{Small Data}}, November 2016.

\bibitem{andreasMeasuringCompositionalityRepresentation2019}
Jacob Andreas.
\newblock Measuring compositionality in representation learning.
\newblock In {\em International Conference on Learning Representations}, 2019.

\bibitem{andreasNeuralModuleNetworks2016}
Jacob Andreas, Marcus Rohrbach, Trevor Darrell, and Dan Klein.
\newblock Neural {{Module Networks}}.
\newblock In {\em 2016 {{IEEE Conference}} on {{Computer Vision}} and {{Pattern
  Recognition}} ({{CVPR}})}, pages 39--48, {Las Vegas, NV, USA}, June 2016.
  {IEEE}.

\bibitem{audaModularNeuralNetworks1999}
G.~Auda and M.~Kamel.
\newblock Modular neural networks a survey.
\newblock {\em International journal of neural systems}, 9 2:129--51, 1999.

\bibitem{avigadModularityMathematics2020}
Jeremy Avigad.
\newblock Modularity in mathematics.
\newblock {\em The Review of Symbolic Logic}, 13(1):47--79, March 2020.

\bibitem{azamBiologicallyInspiredModular2000}
Farooq Azam.
\newblock {\em Biologically {{Inspired Modular Neural Networks}}}.
\newblock PhD thesis, Virginia Tech, May 2000.

\bibitem{bahdanauSystematicGeneralizationWhat2019}
Dzmitry Bahdanau, Shikhar Murty, Michael Noukhovitch, Thien~Huu Nguyen, Harm
  {de Vries}, and Aaron Courville.
\newblock Systematic generalization: {{What}} is required and can it be
  learned?
\newblock In {\em International Conference on Learning Representations}, 2019.

\bibitem{baldwinDesignRulesPower1999}
Carliss~Y. Baldwin and Kim~B. Clark.
\newblock {\em Design {{Rules}}: {{The Power}} of {{Modularity}}}, volume~1.
\newblock {Cambridge, MA: MIT Press}, first edition, 1999.

\bibitem{balestrieroPOLICEProvablyOptimal2022}
Randall Balestriero and Yann LeCun.
\newblock {{POLICE}}: {{Provably Optimal Linear Constraint Enforcement}} for
  {{Deep Neural Networks}}, November 2022.

\bibitem{baltrusaitisMultimodalMachineLearning2018}
Tadas Baltru{\v s}aitis, Chaitanya Ahuja, and Louis-Philippe Morency.
\newblock Multimodal machine learning: {{A}} survey and taxonomy.
\newblock {\em IEEE transactions on pattern analysis and machine intelligence},
  41(2):423--443, 2018.

\bibitem{baoBEiTBERTPreTraining2022}
Hangbo Bao, Li~Dong, Songhao Piao, and Furu Wei.
\newblock {{BEiT}}: {{BERT Pre-Training}} of {{Image Transformers}}.
\newblock In {\em International {{Conference}} on {{Learning
  Representations}}}, 2022.

\bibitem{barhamPathwaysAsynchronousDistributed2022}
Paul Barham, Aakanksha Chowdhery, Jeff Dean, Sanjay Ghemawat, Steven Hand, Dan
  Hurt, Michael Isard, Hyeontaek Lim, Ruoming Pang, Sudip Roy, Brennan Saeta,
  Parker Schuh, Ryan Sepassi, Laurent~El Shafey, Chandramohan~A. Thekkath, and
  Yonghui Wu.
\newblock Pathways: {{Asynchronous Distributed Dataflow}} for {{ML}}.
\newblock {\em arXiv:2203.12533 [cs]}, March 2022.

\bibitem{barrettModularityCognitionFraming2006}
H.~Clark Barrett and Robert Kurzban.
\newblock Modularity in cognition: Framing the debate.
\newblock {\em Psychological Review}, 113(3):628--647, July 2006.

\bibitem{battagliaRelationalInductiveBiases2018}
Peter~W. Battaglia, Jessica~B. Hamrick, Victor Bapst, Alvaro
  {Sanchez-Gonzalez}, Vinicius Zambaldi, Mateusz Malinowski, Andrea Tacchetti,
  David Raposo, Adam Santoro, Ryan Faulkner, Caglar Gulcehre, Francis Song,
  Andrew Ballard, Justin Gilmer, George Dahl, Ashish Vaswani, Kelsey Allen,
  Charles Nash, Victoria Langston, Chris Dyer, Nicolas Heess, Daan Wierstra,
  Pushmeet Kohli, Matt Botvinick, Oriol Vinyals, Yujia Li, and Razvan Pascanu.
\newblock Relational inductive biases, deep learning, and graph networks.
\newblock {\em arXiv:1806.01261 [cs, stat]}, October 2018.

\bibitem{bekkersRotoTranslationCovariantConvolutional2018}
Erik~J. Bekkers, Maxime~W. Lafarge, Mitko Veta, Koen~AJ Eppenhof, Josien~PW
  Pluim, and Remco Duits.
\newblock Roto-{{Translation Covariant Convolutional Networks}} for {{Medical
  Image Analysis}}.
\newblock {\em arXiv:1804.03393 [cs, math]}, June 2018.

\bibitem{belayFactoredConvolutionalNeural2019}
Birhanu Belay, Tewodros Habtegebrial, Marcus Liwicki, Gebeyehu Belay, and
  Didier Stricker.
\newblock Factored {{Convolutional Neural Network}} for {{Amharic Character
  Image Recognition}}.
\newblock In {\em 2019 {{IEEE International Conference}} on {{Image
  Processing}} ({{ICIP}})}, pages 2906--2910, 2019.

\bibitem{benaExtremeSparsityGives2021}
Gabriel B{\'e}na and Dan F.~M. Goodman.
\newblock Extreme sparsity gives rise to functional specialization.
\newblock {\em arXiv:2106.02626 [cs, q-bio]}, June 2021.

\bibitem{bengioConditionalComputationNeural2016}
Emmanuel Bengio, Pierre-Luc Bacon, Joelle Pineau, and Doina Precup.
\newblock Conditional {{Computation}} in {{Neural Networks}} for faster models.
\newblock {\em arXiv:1511.06297 [cs]}, January 2016.

\bibitem{bengioEstimatingPropagatingGradients2013}
Yoshua Bengio, Nicholas L{\'e}onard, and Aaron Courville.
\newblock Estimating or {{Propagating Gradients Through Stochastic Neurons}}
  for {{Conditional Computation}}.
\newblock {\em arXiv:1308.3432 [cs]}, August 2013.

\bibitem{bongardEvolvingModularGenetic2002}
J.~Bongard.
\newblock Evolving modular genetic regulatory networks.
\newblock In {\em Proceedings of the 2002 {{Congress}} on {{Evolutionary
  Computation}}. {{CEC}}'02 ({{Cat}}. {{No}}.{{02TH8600}})}, volume~2, pages
  1872--1877 vol.2, May 2002.

\bibitem{bouchacourtGroundingInductiveBiases2021}
Diane Bouchacourt, Mark Ibrahim, and Ari Morcos.
\newblock Grounding inductive biases in natural images: Invariance stems from
  variations in data.
\newblock In M.~Ranzato, A.~Beygelzimer, Y.~Dauphin, P.S. Liang, and J.~Wortman
  Vaughan, editors, {\em Advances in Neural Information Processing Systems},
  volume~34, pages 19566--19579. {Curran Associates, Inc.}, 2021.

\bibitem{bourbakiArchitectureMathematics1950}
Nicholas Bourbaki.
\newblock The {{Architecture}} of {{Mathematics}}.
\newblock {\em The American Mathematical Monthly}, 57(4):221--232, April 1950.

\bibitem{brandesModularityClustering2007}
Ulrik Brandes, Daniel Delling, Marco Gaertler, Robert Gorke, Martin Hoefer,
  Zoran Nikoloski, and Dorothea Wagner.
\newblock On modularity clustering.
\newblock {\em IEEE transactions on knowledge and data engineering},
  20(2):172--188, 2007.

\bibitem{braylanReuseNeuralModules2016}
Alexander Braylan, Mark Hollenbeck, Elliot Meyerson, and Risto Miikkulainen.
\newblock Reuse of neural modules for general video game playing.
\newblock In {\em Proceedings of the {{AAAI}} Conference on Artificial
  Intelligence}, volume~30, 2016.

\bibitem{bromleySignatureVerificationUsing1994}
Jane Bromley, Isabelle Guyon, Yann LeCun, Eduard S{\"a}ckinger, and Roopak
  Shah.
\newblock Signature {{Verification}} using a "{{Siamese}}" {{Time Delay Neural
  Network}}.
\newblock In J.~Cowan, G.~Tesauro, and J.~Alspector, editors, {\em Advances in
  {{Neural Information Processing Systems}}}, volume~6. {Morgan-Kaufmann},
  1994.

\bibitem{brownLanguageModelsAre2020}
Tom Brown, Benjamin Mann, Nick Ryder, Melanie Subbiah, Jared~D Kaplan, Prafulla
  Dhariwal, Arvind Neelakantan, Pranav Shyam, Girish Sastry, Amanda Askell,
  Sandhini Agarwal, Ariel {Herbert-Voss}, Gretchen Krueger, Tom Henighan, Rewon
  Child, Aditya Ramesh, Daniel Ziegler, Jeffrey Wu, Clemens Winter, Chris
  Hesse, Mark Chen, Eric Sigler, Mateusz Litwin, Scott Gray, Benjamin Chess,
  Jack Clark, Christopher Berner, Sam McCandlish, Alec Radford, Ilya Sutskever,
  and Dario Amodei.
\newblock Language models are few-shot learners.
\newblock In H.~Larochelle, M.~Ranzato, R.~Hadsell, M.F. Balcan, and H.~Lin,
  editors, {\em Advances in Neural Information Processing Systems}, volume~33,
  pages 1877--1901. {Curran Associates, Inc.}, 2020.

\bibitem{caoDeFormerDecomposingPretrained2020}
Qingqing Cao, Harsh Trivedi, Aruna Balasubramanian, and Niranjan
  Balasubramanian.
\newblock {{DeFormer}}: {{Decomposing Pre-trained Transformers}} for {{Faster
  Question Answering}}.
\newblock In {\em Proceedings of the 58th {{Annual Meeting}} of the
  {{Association}} for {{Computational Linguistics}}}, pages 4487--4497,
  {Online}, July 2020. {Association for Computational Linguistics}.

\bibitem{caseySurveyMethodsStrategies1996}
Richard~G Casey and Eric Lecolinet.
\newblock A survey of methods and strategies in character segmentation.
\newblock {\em IEEE transactions on pattern analysis and machine intelligence},
  18(7):690--706, 1996.

\bibitem{changAutomaticallyComposingRepresentation2019}
Michael Chang, Abhishek Gupta, Sergey Levine, and Thomas~L. Griffiths.
\newblock Automatically composing representation transformations as a means for
  generalization.
\newblock In {\em International Conference on Learning Representations}, 2019.

\bibitem{changModularityReinforcementLearning2021}
Michael Chang, Sid Kaushik, Sergey Levine, and Tom Griffiths.
\newblock Modularity in {{Reinforcement Learning}} via {{Algorithmic
  Independence}} in {{Credit Assignment}}.
\newblock In {\em International {{Conference}} on {{Machine Learning}}}, pages
  1452--1462. {PMLR}, July 2021.

\bibitem{chenGrouptheoreticFrameworkData2020}
Shuxiao Chen, Edgar Dobriban, and Jane Lee.
\newblock A group-theoretic framework for data augmentation.
\newblock In H.~Larochelle, M.~Ranzato, R.~Hadsell, M.F. Balcan, and H.~Lin,
  editors, {\em Advances in Neural Information Processing Systems}, volume~33,
  pages 21321--21333. {Curran Associates, Inc.}, 2020.

\bibitem{chenTextRecognitionWild2020}
Xiaoxue Chen, Lianwen Jin, Yuanzhi Zhu, Canjie Luo, and Tianwei Wang.
\newblock Text {{Recognition}} in the {{Wild}}: {{A Survey}}.
\newblock {\em arXiv:2005.03492 [cs]}, December 2020.

\bibitem{chenExploringSimpleSiamese2021}
Xinlei Chen and Kaiming He.
\newblock Exploring {{Simple Siamese Representation Learning}}.
\newblock In {\em 2021 {{IEEE}}/{{CVF Conference}} on {{Computer Vision}} and
  {{Pattern Recognition}} ({{CVPR}})}, pages 15745--15753, {Nashville, TN,
  USA}, June 2021. {IEEE}.

\bibitem{chenModularMetalearningShrinkage2020}
Yutian Chen, Abram~L Friesen, Feryal Behbahani, Arnaud Doucet, David Budden,
  Matthew Hoffman, and Nando {de Freitas}.
\newblock Modular meta-learning with shrinkage.
\newblock {\em Advances in Neural Information Processing Systems},
  33:2858--2869, 2020.

\bibitem{chevalierLongShorttermMemory2022}
Guillaume Chevalier.
\newblock Long short-term memory ({{LSTM}} cell).
\newblock {\em Wikipedia}, September 2022.

\bibitem{choPropertiesNeuralMachine2014}
Kyunghyun Cho, Bart {van Merri{\"e}nboer}, Dzmitry Bahdanau, and Yoshua Bengio.
\newblock On the properties of neural machine translation:
  {{Encoder}}\textendash{{Decoder}} approaches.
\newblock {\em Syntax, Semantics and Structure in Statistical Translation},
  page 103, 2014.

\bibitem{cholletXceptionDeepLearning2017}
Fran{\c c}ois Chollet.
\newblock Xception: {{Deep}} learning with depthwise separable convolutions.
\newblock In {\em Proceedings of the {{IEEE}} Conference on Computer Vision and
  Pattern Recognition}, pages 1251--1258, 2017.

\bibitem{chomskyAspectsTheorySyntax1965}
Noam Chomsky.
\newblock {\em Aspects of the Theory of Syntax}.
\newblock {MIT Press}, 1965.

\bibitem{choudharyNewCharacterSegmentation2013}
Amit Choudhary, Rahul Rishi, and Savita Ahlawat.
\newblock A new character segmentation approach for off-line cursive
  handwritten words.
\newblock {\em Procedia Computer Science}, 17:88--95, 2013.

\bibitem{chowdheryPaLMScalingLanguage2022}
Aakanksha Chowdhery, Sharan Narang, Jacob Devlin, Maarten Bosma, Gaurav Mishra,
  Adam Roberts, Paul Barham, Hyung~Won Chung, Charles Sutton, Sebastian
  Gehrmann, Parker Schuh, Kensen Shi, Sasha Tsvyashchenko, Joshua Maynez,
  Abhishek Rao, Parker Barnes, Yi~Tay, Noam Shazeer, Vinodkumar Prabhakaran,
  Emily Reif, Nan Du, Ben Hutchinson, Reiner Pope, James Bradbury, Jacob
  Austin, Michael Isard, Guy {Gur-Ari}, Pengcheng Yin, Toju Duke, Anselm
  Levskaya, Sanjay Ghemawat, Sunipa Dev, Henryk Michalewski, Xavier Garcia,
  Vedant Misra, Kevin Robinson, Liam Fedus, Denny Zhou, Daphne Ippolito, David
  Luan, Hyeontaek Lim, Barret Zoph, Alexander Spiridonov, Ryan Sepassi, David
  Dohan, Shivani Agrawal, Mark Omernick, Andrew~M. Dai,
  Thanumalayan~Sankaranarayana Pillai, Marie Pellat, Aitor Lewkowycz, Erica
  Moreira, Rewon Child, Oleksandr Polozov, Katherine Lee, Zongwei Zhou, Xuezhi
  Wang, Brennan Saeta, Mark Diaz, Orhan Firat, Michele Catasta, Jason Wei,
  Kathy {Meier-Hellstern}, Douglas Eck, Jeff Dean, Slav Petrov, and Noah
  Fiedel.
\newblock {{PaLM}}: {{Scaling Language Modeling}} with {{Pathways}}.
\newblock {\em arXiv:2204.02311 [cs]}, April 2022.

\bibitem{chuBestPracticesFinetuning2016}
Brian Chu, Vashisht Madhavan, Oscar Beijbom, Judy Hoffman, and Trevor Darrell.
\newblock Best practices for fine-tuning visual classifiers to new domains.
\newblock In {\em European Conference on Computer Vision}, pages 435--442.
  {Springer}, 2016.

\bibitem{claveraPolicyTransferModularity2017}
Ignasi Clavera, David Held, and Pieter Abbeel.
\newblock Policy transfer via modularity and reward guiding.
\newblock In {\em 2017 {{IEEE}}/{{RSJ}} International Conference on Intelligent
  Robots and Systems ({{IROS}})}, pages 1537--1544. {IEEE}, 2017.

\bibitem{cluneEvolutionaryOriginsModularity2013}
Jeff Clune, Jean-Baptiste Mouret, and Hod Lipson.
\newblock The evolutionary origins of modularity.
\newblock {\em Proceedings of the Royal Society b: Biological sciences},
  280(1755):20122863, 2013.

\bibitem{cohenGroupEquivariantConvolutional2016}
Taco~S. Cohen and Max Welling.
\newblock Group {{Equivariant Convolutional Networks}}.
\newblock {\em arXiv:1602.07576 [cs, stat]}, June 2016.

\bibitem{cohenSteerableCNNs2016}
Taco~S. Cohen and Max Welling.
\newblock Steerable {{CNNs}}.
\newblock {\em arXiv:1612.08498 [cs, stat]}, December 2016.

\bibitem{cohen-boulakiaScientificWorkflowsComputational2017}
Sarah {Cohen-Boulakia}, Khalid Belhajjame, Olivier Collin, J{\'e}r{\^o}me
  Chopard, Christine Froidevaux, Alban Gaignard, Konrad Hinsen, Pierre
  Larmande, Yvan Le~Bras, Fr{\'e}d{\'e}ric Lemoine, et~al.
\newblock Scientific workflows for computational reproducibility in the life
  sciences: {{Status}}, challenges and opportunities.
\newblock {\em Future Generation Computer Systems}, 75:284--298, 2017.

\bibitem{colasCuriousIntrinsicallyMotivated2019}
C{\'e}dric Colas, Pierre Fournier, Mohamed Chetouani, Olivier Sigaud, and
  Pierre-Yves Oudeyer.
\newblock Curious: Intrinsically motivated modular multi-goal reinforcement
  learning.
\newblock In {\em International Conference on Machine Learning}, pages
  1331--1340. {PMLR}, 2019.

\bibitem{cormenIntroductionAlgorithms2022}
Thomas~H Cormen, Charles~E Leiserson, Ronald~L Rivest, and Clifford Stein.
\newblock {\em Introduction to Algorithms}.
\newblock 2022.

\bibitem{cosmidesCognitiveAdaptationsSocial1992}
Leda Cosmides and John Tooby.
\newblock Cognitive {{Adaptations}} for {{Social Exchange}}.
\newblock {\em undefined}, pages 163--228, 1992.

\bibitem{cosmidesOriginsDomainSpecificity1994}
Leda Cosmides and John Tooby.
\newblock Origins of domain specificity: {{The}} evolution of functional
  organization.
\newblock In Lawrence~A. Hirschfeld and Susan~A. Gelman, editors, {\em Mapping
  the {{Mind}}: {{Domain Specificity}} in {{Cognition}} and {{Culture}}}, pages
  85--116. {Cambridge University Press}, {Cambridge}, 1994.

\bibitem{csordasCTLEvaluatingGeneralization2022}
R{\'o}bert Csord{\'a}s, Kazuki Irie, and J{\"u}rgen Schmidhuber.
\newblock {{CTL}}++: {{Evaluating Generalization}} on {{Never-Seen
  Compositional Patterns}} of {{Known Functions}}, and {{Compatibility}} of
  {{Neural Representations}}.
\newblock In {\em Proc. {{Conf}}. on {{Empirical Methods}} in {{Natural
  Language Processing}} ({{EMNLP}})}, December 2022.

\bibitem{csordasAreNeuralNets2021}
R{\'o}bert Csord{\'a}s, Sjoerd van Steenkiste, and J{\"u}rgen Schmidhuber.
\newblock Are {{Neural Nets Modular}}? {{Inspecting Functional Modularity
  Through Differentiable Weight Masks}}.
\newblock In {\em International {{Conference}} on {{Learning
  Representations}}}, 2021.

\bibitem{damarioHowModularShould2021}
Vanessa D'Amario, Tomotake Sasaki, and Xavier Boix.
\newblock How {{Modular}} should {{Neural Module Networks Be}} for {{Systematic
  Generalization}}?
\newblock In {\em Thirty-{{Fifth Conference}} on {{Neural Information
  Processing Systems}}}, 2021.

\bibitem{dengImageNetLargescaleHierarchical2009}
Jia Deng, Wei Dong, Richard Socher, Li-Jia Li, Kai Li, and Li~{Fei-Fei}.
\newblock {{ImageNet}}: {{A}} large-scale hierarchical image database.
\newblock In {\em 2009 {{IEEE}} Conference on Computer Vision and Pattern
  Recognition}, pages 248--255, 2009.

\bibitem{devinLearningModularNeural2017}
Coline Devin, Abhishek Gupta, Trevor Darrell, Pieter Abbeel, and Sergey Levine.
\newblock Learning modular neural network policies for multi-task and
  multi-robot transfer.
\newblock In {\em 2017 {{IEEE}} International Conference on Robotics and
  Automation ({{ICRA}})}, pages 2169--2176. {IEEE}, 2017.

\bibitem{dielemanExploitingCyclicSymmetry2016}
Sander Dieleman, Jeffrey De~Fauw, and Koray Kavukcuoglu.
\newblock Exploiting {{Cyclic Symmetry}} in {{Convolutional Neural Networks}}.
\newblock {\em arXiv:1602.02660 [cs]}, May 2016.

\bibitem{dingTrunkbranchEnsembleConvolutional2017}
Changxing Ding and Dacheng Tao.
\newblock Trunk-branch ensemble convolutional neural networks for video-based
  face recognition.
\newblock {\em IEEE transactions on pattern analysis and machine intelligence},
  40(4):1002--1014, 2017.

\bibitem{dosovitskiyImageWorth16x162021}
Alexey Dosovitskiy, Lucas Beyer, Alexander Kolesnikov, Dirk Weissenborn,
  Xiaohua Zhai, Thomas Unterthiner, Mostafa Dehghani, Matthias Minderer, Georg
  Heigold, Sylvain Gelly, Jakob Uszkoreit, and Neil Houlsby.
\newblock An {{Image}} is {{Worth}} 16x16 {{Words}}: {{Transformers}} for
  {{Image Recognition}} at {{Scale}}.
\newblock In {\em International {{Conference}} on {{Learning
  Representations}}}, 2021.

\bibitem{duGlamEfficientScaling2022}
Nan Du, Yanping Huang, Andrew~M Dai, Simon Tong, Dmitry Lepikhin, Yuanzhong Xu,
  Maxim Krikun, Yanqi Zhou, Adams~Wei Yu, Orhan Firat, et~al.
\newblock Glam: {{Efficient}} scaling of language models with
  mixture-of-experts.
\newblock In {\em International Conference on Machine Learning}, pages
  5547--5569. {PMLR}, 2022.

\bibitem{eastwoodFrameworkQuantitativeEvaluation2018}
Cian Eastwood and Christopher K.~I. Williams.
\newblock A {{Framework}} for the {{Quantitative Evaluation}} of {{Disentangled
  Representations}}.
\newblock In {\em Sixth {{International Conference}} on {{Learning
  Representations}} ({{ICLR}} 2018)}, May 2018.

\bibitem{eigenLearningFactoredRepresentations2014}
David Eigen, Marc'Aurelio Ranzato, and Ilya Sutskever.
\newblock Learning factored representations in a deep mixture of experts.
\newblock In {\em {{ICLR Workshop}}}, 2014.

\bibitem{elbazLessonsLearnedNeurIPS2022a}
Adrian El~Baz, Ihsan Ullah, Edesio Alcoba{\c c}a, Andr{\'e} C. P. L.~F.
  Carvalho, Hong Chen, Fabio Ferreira, Henry Gouk, Chaoyu Guan, Isabelle Guyon,
  Timothy Hospedales, Shell Hu, Mike Huisman, Frank Hutter, Zhengying Liu,
  Felix Mohr, Ekrem {\"O}zt{\"u}rk, Jan~N. {van Rijn}, Haozhe Sun, Xin Wang,
  and Wenwu Zhu.
\newblock Lessons learned from the {{NeurIPS}} 2021 {{MetaDL}} challenge:
  {{Backbone}} fine-tuning without episodic meta-learning dominates for
  few-shot learning image classification.
\newblock In Douwe Kiela, Marco Ciccone, and Barbara Caputo, editors, {\em
  Proceedings of the {{NeurIPS}} 2021 Competitions and Demonstrations Track},
  volume 176 of {\em Proceedings of Machine Learning Research}, pages 80--96.
  {PMLR}, December 2022.

\bibitem{ellefsenNeuralModularityHelps2015}
Kai~Olav Ellefsen, Jean-Baptiste Mouret, and Jeff Clune.
\newblock Neural {{Modularity Helps Organisms Evolve}} to {{Learn New Skills}}
  without {{Forgetting Old Skills}}.
\newblock {\em PLOS Computational Biology}, 11(4):e1004128, April 2015.

\bibitem{elsayedRevisitingSpatialInvariance2020}
Gamaleldin~F. Elsayed, Prajit Ramachandran, Jonathon Shlens, and Simon
  Kornblith.
\newblock Revisiting {{Spatial Invariance}} with {{Low-Rank Local
  Connectivity}}.
\newblock {\em arXiv:2002.02959 [cs, stat]}, August 2020.

\bibitem{elskenNeuralArchitectureSearch}
Thomas Elsken, Jan~Hendrik Metzen, and Frank Hutter.
\newblock Neural {{Architecture Search}}.
\newblock pages 69--86.

\bibitem{fedusReviewSparseExpert2022}
William Fedus, Jeff Dean, and Barret Zoph.
\newblock A {{Review}} of {{Sparse Expert Models}} in {{Deep Learning}},
  September 2022.

\bibitem{fedusSwitchTransformersScaling2022}
William Fedus, Barret Zoph, and Noam Shazeer.
\newblock Switch transformers: {{Scaling}} to trillion parameter models with
  simple and efficient sparsity.
\newblock {\em Journal of Machine Learning Research}, 23(120):1--39, 2022.

\bibitem{fernandoPathNetEvolutionChannels2017a}
Chrisantha Fernando, Dylan Banarse, Charles Blundell, Yori Zwols, David Ha,
  Andrei~A. Rusu, Alexander Pritzel, and Daan Wierstra.
\newblock {{PathNet}}: {{Evolution Channels Gradient Descent}} in {{Super
  Neural Networks}}.
\newblock {\em arXiv:1701.08734 [cs]}, January 2017.

\bibitem{filanClusterabilityNeuralNetworks2021}
Daniel Filan, Stephen Casper, Shlomi Hod, Cody Wild, Andrew Critch, and Stuart
  Russell.
\newblock Clusterability in {{Neural Networks}}.
\newblock {\em arXiv:2103.03386 [cs]}, March 2021.

\bibitem{finnModelAgnosticMetaLearningFast2017}
Chelsea Finn, Pieter Abbeel, and Sergey Levine.
\newblock Model-{{Agnostic Meta-Learning}} for {{Fast Adaptation}} of {{Deep
  Networks}}.
\newblock {\em arXiv:1703.03400 [cs]}, July 2017.

\bibitem{fodorModularityMind1983}
Jerry~A. Fodor.
\newblock The {{Modularity}} of {{Mind}}.
\newblock April 1983.

\bibitem{fodorMindDoesnWork2000}
Jerry~A. Fodor.
\newblock {\em The Mind Doesn't Work That Way: {{The}} Scope and Limits of
  Computational Psychology}.
\newblock {MIT Press}, 2000.

\bibitem{fodorConnectionismCognitiveArchitecture1988}
Jerry~A Fodor and Zenon~W Pylyshyn.
\newblock Connectionism and cognitive architecture: {{A}} critical analysis.
\newblock {\em Cognition}, 28(1-2):3--71, 1988.

\bibitem{fordArchitectsIntelligenceTruth2018}
Martin Ford.
\newblock {\em Architects of Intelligence: The Truth about {{AI}} from the
  People Building It}.
\newblock {Packt Publishing}, {Birmingham, UK}, first published: november 2018
  edition, 2018.

\bibitem{frankenhuisEvolutionaryPsychologyFodor2007}
Willem~E. Frankenhuis and Annemie Ploeger.
\newblock Evolutionary {{Psychology Versus Fodor}}: {{Arguments For}} and
  {{Against}} the {{Massive Modularity Hypothesis}}.
\newblock {\em Philosophical Psychology}, 20(6):687--710, December 2007.

\bibitem{frenchUsingSemiDistributedRepresentations1991}
Robert French.
\newblock Using {{Semi-Distributed Representations}} to {{Overcome Catastrophic
  Forgetting}} in {{Connectionist Networks}}.
\newblock 1991.

\bibitem{fuengfusinNetworkSubnetworksLayerwise2020}
Ninnart Fuengfusin and Hakaru Tamukoh.
\newblock Network with {{Sub-networks}}: {{Layer-wise Detachable Neural
  Network}}.
\newblock {\em Journal of Robotics, Networking and Artificial Life},
  7(4):240--244, 2020.

\bibitem{galantiModularityHypernetworks2020}
Tomer Galanti and Lior Wolf.
\newblock On the {{Modularity}} of {{Hypernetworks}}.
\newblock {\em arXiv:2002.10006 [cs, stat]}, November 2020.

\bibitem{gaoEfficientInvariantConvolutional2017}
Hongyang Gao and Shuiwang Ji.
\newblock Efficient and {{Invariant Convolutional Neural Networks}} for {{Dense
  Prediction}}.
\newblock In {\em 2017 {{IEEE International Conference}} on {{Data Mining}}
  ({{ICDM}})}, pages 871--876, 2017.

\bibitem{gatysImageStyleTransfer2016}
Leon~A. Gatys, Alexander~S. Ecker, and Matthias Bethge.
\newblock Image {{Style Transfer Using Convolutional Neural Networks}}.
\newblock In {\em 2016 {{IEEE Conference}} on {{Computer Vision}} and {{Pattern
  Recognition}} ({{CVPR}})}, pages 2414--2423, {Las Vegas, NV, USA}, June 2016.
  {IEEE}.

\bibitem{gavaliChapterDeepConvolutional2019}
Pralhad Gavali and J.~Saira Banu.
\newblock Chapter 6 - deep convolutional neural network for image
  classification on {{CUDA}} platform.
\newblock In Arun~Kumar Sangaiah, editor, {\em Deep Learning and Parallel
  Computing Environment for Bioengineering Systems}, pages 99--122. {Academic
  Press}, 2019.

\bibitem{gentileTheoryModularityHypothesis2013}
Peter Gentile.
\newblock Theory of {{Modularity}}, a {{Hypothesis}}.
\newblock {\em Procedia Computer Science}, 20, December 2013.

\bibitem{ghaziRecursiveSketchesModular2019}
Badih Ghazi, Rina Panigrahy, and Joshua Wang.
\newblock Recursive {{Sketches}} for {{Modular Deep Learning}}.
\newblock In {\em Proceedings of the 36th {{International Conference}} on
  {{Machine Learning}}}, pages 2211--2220. {PMLR}, May 2019.

\bibitem{gomezNewModularityMeasure2016}
Daniel G{\'o}mez, J.~Tinguaro~Rodr{\'i}guez, Javier Y{\'a}{\~n}ez, and Javier
  Montero.
\newblock A new modularity measure for {{Fuzzy Community}} detection problems
  based on overlap and grouping functions.
\newblock {\em International Journal of Approximate Reasoning}, 74:88--107,
  July 2016.

\bibitem{goodfellowDeepLearning2016}
Ian Goodfellow, Yoshua Bengio, and Aaron Courville.
\newblock {\em Deep {{Learning}}}.
\newblock {MIT Press}, 2016.

\bibitem{goodfellowGenerativeAdversarialNetworks2014}
Ian~J. Goodfellow, Jean {Pouget-Abadie}, Mehdi Mirza, Bing Xu, David
  {Warde-Farley}, Sherjil Ozair, Aaron Courville, and Yoshua Bengio.
\newblock Generative {{Adversarial Networks}}.
\newblock {\em arXiv:1406.2661 [cs, stat]}, June 2014.

\bibitem{goyalRecurrentIndependentMechanisms2021}
Anirudh Goyal, Alex Lamb, Jordan Hoffmann, Shagun Sodhani, Sergey Levine,
  Yoshua Bengio, and Bernhard Sch{\"o}lkopf.
\newblock Recurrent independent mechanisms.
\newblock In {\em International Conference on Learning Representations}, 2021.

\bibitem{goyalMakingVqaMatter2017}
Yash Goyal, Tejas Khot, Douglas {Summers-Stay}, Dhruv Batra, and Devi Parikh.
\newblock Making the v in vqa matter: {{Elevating}} the role of image
  understanding in visual question answering.
\newblock In {\em Proceedings of the {{IEEE}} Conference on Computer Vision and
  Pattern Recognition}, pages 6904--6913, 2017.

\bibitem{grayGPUKernelsBlockSparse}
Scott Gray, Alec Radford, and Diederik~P Kingma.
\newblock {{GPU Kernels}} for {{Block-Sparse Weights}}.
\newblock Technical report.

\bibitem{haHyperNetworks2016}
David Ha, Andrew Dai, and Quoc~V. Le.
\newblock {{HyperNetworks}}.
\newblock {\em arXiv:1609.09106 [cs]}, December 2016.

\bibitem{hacohenPowerCurriculumLearning2019}
Guy Hacohen and Daphna Weinshall.
\newblock On {{The Power}} of {{Curriculum Learning}} in {{Training Deep
  Networks}}.
\newblock In Kamalika Chaudhuri and Ruslan Salakhutdinov, editors, {\em
  Proceedings of the 36th {{International Conference}} on {{Machine
  Learning}}}, volume~97 of {\em Proceedings of {{Machine Learning Research}}},
  pages 2535--2544. {PMLR}, June 2019.

\bibitem{hastieElementsStatisticalLearning2009}
Trevor Hastie, Robert Tibshirani, and Jerome Friedman.
\newblock {\em The Elements of Statistical Learning: Data Mining, Inference and
  Prediction}.
\newblock {Springer}, second edition, 2009.

\bibitem{heFasterMoEModelingOptimizing2022}
Jiaao He, Jidong Zhai, Tiago Antunes, Haojie Wang, Fuwen Luo, Shangfeng Shi,
  and Qin Li.
\newblock {{FasterMoE}}: Modeling and optimizing training of large-scale
  dynamic pre-trained models.
\newblock In {\em Proceedings of the 27th {{ACM SIGPLAN}} Symposium on
  Principles and Practice of Parallel Programming}, pages 120--134, 2022.

\bibitem{heMaskedAutoencodersAre2022}
Kaiming He, Xinlei Chen, Saining Xie, Yanghao Li, Piotr Doll{\'a}r, and Ross
  Girshick.
\newblock Masked autoencoders are scalable vision learners.
\newblock In {\em Proceedings of the {{IEEE}}/{{CVF Conference}} on {{Computer
  Vision}} and {{Pattern Recognition}}}, pages 16000--16009, 2022.

\bibitem{heRethinkingImagenetPretraining2019}
Kaiming He, Ross Girshick, and Piotr Doll{\'a}r.
\newblock Rethinking imagenet pre-training.
\newblock In {\em Proceedings of the {{IEEE}}/{{CVF}} International Conference
  on Computer Vision}, pages 4918--4927, 2019.

\bibitem{heDeepResidualLearning2016}
Kaiming He, Xiangyu Zhang, Shaoqing Ren, and Jian Sun.
\newblock Deep residual learning for image recognition.
\newblock In {\em Proceedings of the {{IEEE}} Conference on Computer Vision and
  Pattern Recognition}, pages 770--778, 2016.

\bibitem{hintonReducingDimensionalityData2006}
G.~E. Hinton and R.~R. Salakhutdinov.
\newblock Reducing the {{Dimensionality}} of {{Data}} with {{Neural Networks}}.
\newblock {\em Science}, 313(5786):504--507, 2006.

\bibitem{hochreiterLongShorttermMemory1997}
Sepp Hochreiter and J{\"u}rgen Schmidhuber.
\newblock Long short-term memory.
\newblock {\em Neural computation}, 9(8):1735--1780, 1997.

\bibitem{hofmanEvolutionHumanBrain2014}
Michel~A Hofman.
\newblock Evolution of the human brain: When bigger is better.
\newblock {\em Frontiers in neuroanatomy}, 8:15, 2014.

\bibitem{howardMobilenetsEfficientConvolutional2017}
Andrew~G Howard, Menglong Zhu, Bo~Chen, Dmitry Kalenichenko, Weijun Wang,
  Tobias Weyand, Marco Andreetto, and Hartwig Adam.
\newblock Mobilenets: {{Efficient}} convolutional neural networks for mobile
  vision applications.
\newblock {\em arXiv preprint arXiv:1704.04861}, 2017.

\bibitem{huDeepStockRepresentation2018}
Guosheng Hu, Yuxin Hu, Kai Yang, Zehao Yu, Flood Sung, Zhihong Zhang, Fei Xie,
  Jianguo Liu, Neil Robertson, Timothy Hospedales, and Qiangwei Miemie.
\newblock Deep {{Stock Representation Learning}}: {{From Candlestick Charts}}
  to {{Investment Decisions}}.
\newblock {\em arXiv:1709.03803 [q-fin]}, February 2018.

\bibitem{huLearningReasonEndtoEnd2017}
Ronghang Hu, Jacob Andreas, Marcus Rohrbach, Trevor Darrell, and Kate Saenko.
\newblock Learning to {{Reason}}: {{End-to-End Module Networks}} for {{Visual
  Question Answering}}.
\newblock In {\em Proceedings of the {{IEEE International Conference}} on
  {{Computer Vision}} ({{ICCV}})}, 2017.

\bibitem{huangMultiplexedNetworkEndtoend2021}
Jing Huang, Guan Pang, Rama Kovvuri, Mandy Toh, Kevin~J Liang, Praveen
  Krishnan, Xi~Yin, and Tal Hassner.
\newblock A multiplexed network for end-to-end, multilingual {{OCR}}.
\newblock In {\em Proceedings of the {{IEEE}}/{{CVF}} Conference on Computer
  Vision and Pattern Recognition}, pages 4547--4557, 2021.

\bibitem{huizingaEvolvingNeuralNetworks2014}
Joost Huizinga, Jeff Clune, and Jean-Baptiste Mouret.
\newblock Evolving neural networks that are both modular and regular:
  {{Hyperneat}} plus the connection cost technique.
\newblock In {\em Proceedings of the 2014 Annual Conference on Genetic and
  Evolutionary Computation}, pages 697--704, 2014.

\bibitem{hupkesCompositionalityDecomposedHow2020}
Dieuwke Hupkes, Verna Dankers, Mathijs Mul, and Elia Bruni.
\newblock Compositionality decomposed: {{How}} do neural networks generalise?
\newblock {\em Journal of Artificial Intelligence Research}, 67:757--795, 2020.

\bibitem{hupkesStateoftheartGeneralisationResearch2022}
Dieuwke Hupkes, Mario Giulianelli, Verna Dankers, Mikel Artetxe, Yanai Elazar,
  Tiago Pimentel, Christos Christodoulopoulos, Karim Lasri, Naomi Saphra,
  Arabella Sinclair, Dennis Ulmer, Florian Schottmann, Khuyagbaatar Batsuren,
  Kaiser Sun, Koustuv Sinha, Leila Khalatbari, Maria Ryskina, Rita Frieske,
  Ryan Cotterell, and Zhijing Jin.
\newblock State-of-the-art generalisation research in {{NLP}}: A taxonomy and
  review, October 2022.

\bibitem{hutterAutomaticMachineLearning2019}
Frank Hutter, Lars Kotthoff, and Joaquin Vanschoren, editors.
\newblock {\em Automatic Machine Learning: {{Methods}}, Systems, Challenges}.
\newblock {Springer}, 2019.

\bibitem{islamDiscreteFactorialRepresentations2022}
Riashat Islam, Hongyu Zang, Anirudh Goyal, Alex Lamb, Kenji Kawaguchi, Xin Li,
  Romain Laroche, Yoshua Bengio, and Remi Tachet~Des Combes.
\newblock Discrete {{Factorial Representations}} as an {{Abstraction}} for
  {{Goal Conditioned Reinforcement Learning}}, October 2022.

\bibitem{jacobsTaskDecompositionCompetition1991}
Robert~A Jacobs, Michael~I Jordan, and Andrew~G Barto.
\newblock Task decomposition through competition in a modular connectionist
  architecture: {{The}} what and where vision tasks.
\newblock {\em Cognitive science}, 15(2):219--250, 1991.

\bibitem{jacobsAdaptiveMixturesLocal1991}
Robert~A. Jacobs, Michael~I. Jordan, Steven~J. Nowlan, and Geoffrey~E. Hinton.
\newblock Adaptive {{Mixtures}} of {{Local Experts}}.
\newblock {\em Neural Computation}, 3(1):79--87, March 1991.

\bibitem{javedMetaLearningRepresentationsContinual2019}
Khurram Javed and Martha White.
\newblock Meta-{{Learning Representations}} for {{Continual Learning}}.
\newblock {\em arXiv:1905.12588 [cs, stat]}, October 2019.

\bibitem{jinSplitCNNSplittingWindowBased2019}
Tian Jin and Seokin Hong.
\newblock Split-{{CNN}}: {{Splitting Window-Based Operations}} in
  {{Convolutional Neural Networks}} for {{Memory System Optimization}}.
\newblock In {\em Proceedings of the {{Twenty-Fourth International Conference}}
  on {{Architectural Support}} for {{Programming Languages}} and {{Operating
  Systems}}}, {{ASPLOS}} '19, pages 835--847, {New York, NY, USA}, 2019.
  {Association for Computing Machinery}.

\bibitem{jingMaskedSiameseConvNets2022}
Li~Jing, Jiachen Zhu, and Yann LeCun.
\newblock Masked {{Siamese ConvNets}}, June 2022.

\bibitem{jordanHierarchicalMixturesExperts1994}
Michael~I Jordan and Robert~A Jacobs.
\newblock Hierarchical mixtures of experts and the {{EM}} algorithm.
\newblock {\em Neural computation}, 6(2):181--214, 1994.

\bibitem{juRelativePerformanceEnsemble2018}
Cheng Ju, Aur{\'e}lien Bibaut, and Mark {van der Laan}.
\newblock The relative performance of ensemble methods with deep convolutional
  neural networks for image classification.
\newblock {\em Journal of Applied Statistics}, 45(15):2800--2818, 2018.

\bibitem{jurafskySpeechLanguageProcessing2019}
Dan Jurafsky and James~H Martin.
\newblock Speech and language processing (3rd draft ed.), 2019.

\bibitem{kanakisReparameterizingConvolutionsIncremental2020}
Menelaos Kanakis, David Bruggemann, Suman Saha, Stamatios Georgoulis, Anton
  Obukhov, and Luc~Van Gool.
\newblock Reparameterizing convolutions for incremental multi-task learning
  without task interference.
\newblock In {\em European {{Conference}} on {{Computer Vision}}}, pages
  689--707. {Springer}, 2020.

\bibitem{kassnerBeliefBankAddingMemory2021}
Nora Kassner, Oyvind Tafjord, Hinrich Sch{\"u}tze, and Peter Clark.
\newblock {{BeliefBank}}: {{Adding Memory}} to a {{Pre-Trained Language Model}}
  for a {{Systematic Notion}} of {{Belief}}.
\newblock In {\em Proceedings of the 2021 {{Conference}} on {{Empirical
  Methods}} in {{Natural Language Processing}}}, pages 8849--8861, 2021.

\bibitem{kaurStudyVariousCharacter2015}
Amandeep Kaur, Seema Baghla, and Sunil Kumar.
\newblock Study of various character segmentation techniques for handwritten
  off-line cursive words: {{A}} review.
\newblock {\em International Journal of Advances in Science Engineering and
  Technology}, 3(3):154--158, 2015.

\bibitem{keAchievingForgettingPrevention2021}
Zixuan Ke, Bing Liu, Nianzu Ma, Hu~Xu, and Lei Shu.
\newblock Achieving forgetting prevention and knowledge transfer in continual
  learning.
\newblock {\em Advances in Neural Information Processing Systems},
  34:22443--22456, 2021.

\bibitem{kentonBERTPretrainingDeep2019}
Jacob Devlin Ming-Wei~Chang Kenton and Lee~Kristina Toutanova.
\newblock {{BERT}}: {{Pre-training}} of deep bidirectional transformers for
  language understanding.
\newblock In {\em Proceedings of {{NAACL-HLT}}}, pages 4171--4186, 2019.

\bibitem{keskarLargeBatchTrainingDeep2017}
Nitish~Shirish Keskar, Dheevatsa Mudigere, Jorge Nocedal, Mikhail Smelyanskiy,
  and Ping Tak~Peter Tang.
\newblock On {{Large-Batch Training}} for {{Deep Learning}}: {{Generalization
  Gap}} and {{Sharp Minima}}.
\newblock In {\em {{ICLR}}}, 2017.

\bibitem{keysersMeasuringCompositionalGeneralization2020}
Daniel Keysers, Nathanael Sch{\"a}rli, Nathan Scales, Hylke Buisman, Daniel
  Furrer, Sergii Kashubin, Nikola Momchev, Danila Sinopalnikov, Lukasz
  Stafiniak, Tibor Tihon, Dmitry Tsarkov, Xiao Wang, Marc {van Zee}, and
  Olivier Bousquet.
\newblock Measuring compositional generalization: {{A}} comprehensive method on
  realistic data.
\newblock In {\em International Conference on Learning Representations}, 2020.

\bibitem{kimSplitNetLearningSemantically2017}
Juyong Kim, Yookoon Park, Gunhee Kim, and Sung~Ju Hwang.
\newblock {{SplitNet}}: {{Learning}} to {{Semantically Split Deep Networks}}
  for {{Parameter Reduction}} and {{Model Parallelization}}.
\newblock In {\em Proceedings of the 34th {{International Conference}} on
  {{Machine Learning}}}, pages 1866--1874. {PMLR}, July 2017.

\bibitem{kingetsuNeuralNetworkModule2021}
Hiroaki Kingetsu, Kenichi Kobayashi, and Taiji Suzuki.
\newblock Neural {{Network Module Decomposition}} and {{Recomposition}},
  December 2021.

\bibitem{kirschModularNetworksLearning2018}
Louis Kirsch, Julius Kunze, and David Barber.
\newblock Modular {{Networks}}: {{Learning}} to {{Decompose Neural
  Computation}}.
\newblock In {\em Advances in {{Neural Information Processing Systems}}},
  volume~31. {Curran Associates, Inc.}, 2018.

\bibitem{kohComparisonAnalysisDeep2021}
Eunjeong Koh and Shlomo Dubnov.
\newblock Comparison and {{Analysis}} of {{Deep Audio Embeddings}} for {{Music
  Emotion Recognition}}, April 2021.

\bibitem{krishnamurthyInterpretabilityGatedModular2021}
Yamuna Krishnamurthy and Chris Watkins.
\newblock Interpretability in gated modular neural networks.
\newblock In {\em {{eXplainable AI}} Approaches for Debugging and Diagnosis.},
  2021.

\bibitem{krizhevskyImageNetClassificationDeep2012}
Alex Krizhevsky, Ilya Sutskever, and Geoffrey~E Hinton.
\newblock {{ImageNet Classification}} with {{Deep Convolutional Neural
  Networks}}.
\newblock In {\em Advances in Neural Information Processing Systems},
  volume~25, 2012.

\bibitem{kruegerZoneoutRegularizingRNNs2017}
David Krueger, Tegan Maharaj, Janos Kramar, Mohammad Pezeshki, Nicolas Ballas,
  Nan~Rosemary Ke, Anirudh Goyal, Yoshua Bengio, Aaron Courville, and
  Christopher Pal.
\newblock Zoneout: {{Regularizing RNNs}} by randomly preserving hidden
  activations.
\newblock In {\em International Conference on Learning Representations}, 2017.

\bibitem{kurzweilHowCreateMind2013}
Ray Kurzweil.
\newblock {\em How to {{Create}} a {{Mind}}: {{The Secret}} of {{Human Thought
  Revealed}}}.
\newblock {Penguin Books}, {USA}, 2013.

\bibitem{laenenEpisodesPrototypicalNetworks2021}
Steinar Laenen and Luca Bertinetto.
\newblock On episodes, prototypical networks, and few-shot learning.
\newblock In M.~Ranzato, A.~Beygelzimer, Y.~Dauphin, P.S. Liang, and J.~Wortman
  Vaughan, editors, {\em Advances in Neural Information Processing Systems},
  volume~34, pages 24581--24592. {Curran Associates, Inc.}, 2021.

\bibitem{lakeHumanlevelConceptLearning2015}
B.~M. Lake, R.~Salakhutdinov, and J.~B. Tenenbaum.
\newblock Human-level concept learning through probabilistic program induction.
\newblock {\em Science}, 350(6266):1332--1338, December 2015.

\bibitem{lakeGeneralizationSystematicityCompositional2018}
Brenden Lake and Marco Baroni.
\newblock Generalization without systematicity: {{On}} the compositional skills
  of sequence-to-sequence recurrent networks.
\newblock In {\em International Conference on Machine Learning}, pages
  2873--2882. {PMLR}, 2018.

\bibitem{lakeCompositionalGeneralizationMeta2019}
Brenden~M. Lake.
\newblock Compositional generalization through meta sequence-to-sequence
  learning.
\newblock {\em arXiv:1906.05381 [cs]}, October 2019.

\bibitem{lecunLearningMethodsGeneric2004}
Y.~LeCun, Fu~Jie Huang, and L.~Bottou.
\newblock Learning methods for generic object recognition with invariance to
  pose and lighting.
\newblock In {\em Proceedings of the 2004 {{IEEE Computer Society Conference}}
  on {{Computer Vision}} and {{Pattern Recognition}}, 2004. {{CVPR}} 2004.},
  volume~2, pages II--104 Vol.2, June 2004.

\bibitem{lecunPathAutonomousMachine2022}
Yann LeCun.
\newblock A path towards autonomous machine intelligence version 0.9. 2,
  2022-06-27.
\newblock 2022.

\bibitem{lecunGradientbasedLearningApplied1998}
Yann LeCun, L{\'e}on Bottou, Yoshua Bengio, and Patrick Haffner.
\newblock Gradient-based learning applied to document recognition.
\newblock {\em Proceedings of the IEEE}, 86(11):2278--2324, 1998.

\bibitem{lecunOptimalBrainDamage1989}
Yann LeCun, John Denker, and Sara Solla.
\newblock Optimal brain damage.
\newblock {\em Advances in neural information processing systems}, 2, 1989.

\bibitem{liIntroductionKolmogorovComplexity2008}
Ming Li and Paul~M.B. Vitnyi.
\newblock {\em An {{Introduction}} to {{Kolmogorov Complexity}} and {{Its
  Applications}}}.
\newblock {Springer Publishing Company, Incorporated}, third edition, 2008.

\bibitem{liNeuralSpeechSynthesis2019}
Naihan Li, Shujie Liu, Yanqing Liu, Sheng Zhao, and Ming Liu.
\newblock Neural speech synthesis with transformer network.
\newblock In {\em Proceedings of the {{AAAI}} Conference on Artificial
  Intelligence}, volume~33, pages 6706--6713, 2019.

\bibitem{liLearningCompositionalVisual2020a}
Zhi Li, Bo~Wu, Qi~Liu, Likang Wu, Hongke Zhao, and Tao Mei.
\newblock Learning the compositional visual coherence for complementary
  recommendations.
\newblock In Christian Bessiere, editor, {\em Proceedings of the Twenty-Ninth
  International Joint Conference on Artificial Intelligence, {{IJCAI-20}}},
  pages 3536--3543. {International Joint Conferences on Artificial Intelligence
  Organization}, July 2020.

\bibitem{liuHierarchicalRepresentationsEfficient2018}
Hanxiao Liu, Karen Simonyan, Oriol Vinyals, Chrisantha Fernando, and Koray
  Kavukcuoglu.
\newblock Hierarchical representations for efficient architecture search.
\newblock In {\em International Conference on Learning Representations}, 2018.

\bibitem{loulaRearrangingFamiliarTesting2018}
Jo{\~a}o Loula, Marco Baroni, and Brenden~M. Lake.
\newblock Rearranging the familiar: {{Testing}} compositional generalization in
  recurrent networks.
\newblock In {\em {{BlackboxNLP}}@{{EMNLP}}}, pages 108--114, 2018.

\bibitem{maDisentangledGraphConvolutional2019}
Jianxin Ma, Peng Cui, Kun Kuang, Xin Wang, and Wenwu Zhu.
\newblock Disentangled graph convolutional networks.
\newblock In Kamalika Chaudhuri and Ruslan Salakhutdinov, editors, {\em
  Proceedings of the 36th International Conference on Machine Learning},
  volume~97 of {\em Proceedings of Machine Learning Research}, pages
  4212--4221. {PMLR}, June 2019.

\bibitem{maninisAttentiveSingleTaskingMultiple2019}
Kevis-Kokitsi Maninis, Ilija Radosavovic, and Iasonas Kokkinos.
\newblock Attentive {{Single-Tasking}} of {{Multiple Tasks}}.
\newblock In {\em 2019 {{IEEE}}/{{CVF Conference}} on {{Computer Vision}} and
  {{Pattern Recognition}} ({{CVPR}})}, pages 1851--1860, {Long Beach, CA, USA},
  June 2019. {IEEE}.

\bibitem{martinabadiTensorFlowLargeScaleMachine2015}
{Mart\'in Abadi}, {Ashish Agarwal}, {Paul Barham}, {Eugene Brevdo}, {Zhifeng
  Chen}, {Craig Citro}, {Greg S. Corrado}, {Andy Davis}, {Jeffrey Dean},
  {Matthieu Devin}, {Sanjay Ghemawat}, {Ian Goodfellow}, {Andrew Harp},
  {Geoffrey Irving}, {Michael Isard}, Yangqing Jia, {Rafal Jozefowicz}, {Lukasz
  Kaiser}, {Manjunath Kudlur}, {Josh Levenberg}, {Dandelion Man\'e}, {Rajat
  Monga}, {Sherry Moore}, {Derek Murray}, {Chris Olah}, {Mike Schuster},
  {Jonathon Shlens}, {Benoit Steiner}, {Ilya Sutskever}, {Kunal Talwar}, {Paul
  Tucker}, {Vincent Vanhoucke}, {Vijay Vasudevan}, {Fernanda Vi\'egas}, {Oriol
  Vinyals}, {Pete Warden}, {Martin Wattenberg}, {Martin Wicke}, {Yuan Yu}, and
  {Xiaoqiang Zheng}.
\newblock {{TensorFlow}}: {{Large-Scale Machine Learning}} on {{Heterogeneous
  Systems}}, 2015.

\bibitem{masseAlleviatingCatastrophicForgetting2018}
Nicolas~Y. Masse, Gregory~D. Grant, and David~J. Freedman.
\newblock Alleviating catastrophic forgetting using context-dependent gating
  and synaptic stabilization.
\newblock {\em Proceedings of the National Academy of Sciences},
  115(44):E10467--E10475, 2018.

\bibitem{mazziaEfficientcapsnetCapsuleNetwork2021}
Vittorio Mazzia, Francesco Salvetti, and Marcello Chiaberge.
\newblock Efficient-capsnet: {{Capsule}} network with self-attention routing.
\newblock {\em Scientific reports}, 11(1):1--13, 2021.

\bibitem{mcneely-whiteInceptionResNetFeatures2020}
David {McNeely-White}, J.~Ross Beveridge, and Bruce~A. Draper.
\newblock Inception and {{ResNet}} features are (almost) equivalent.
\newblock {\em Cognitive Systems Research}, 59:312--318, January 2020.

\bibitem{mengLocatingEditingFactual2022}
Kevin Meng, David Bau, Alex Andonian, and Yonatan Belinkov.
\newblock Locating and {{Editing Factual Associations}} in {{GPT}}.
\newblock February 2022.

\bibitem{meyersonModularUniversalReparameterization2019}
Elliot Meyerson and Risto Miikkulainen.
\newblock Modular universal reparameterization: {{Deep}} multi-task learning
  across diverse domains.
\newblock {\em Advances in Neural Information Processing Systems}, 32, 2019.

\bibitem{mitchellFastModelEditing2021}
Eric Mitchell, Charles Lin, Antoine Bosselut, Chelsea Finn, and Christopher~D.
  Manning.
\newblock Fast {{Model Editing}} at {{Scale}}.
\newblock {\em arXiv:2110.11309 [cs]}, October 2021.

\bibitem{mitchellMemoryBasedModelEditing2022}
Eric Mitchell, Charles Lin, Antoine Bosselut, Chelsea Finn, and Christopher~D.
  Manning.
\newblock Memory-{{Based Model Editing}} at {{Scale}}.
\newblock In {\em International {{Conference}} on {{Machine Learning}}}, 2022.

\bibitem{mittalModularArchitectureEnough2022}
Sarthak Mittal, Yoshua Bengio, and Guillaume Lajoie.
\newblock Is a {{Modular Architecture Enough}}?, 2022.

\bibitem{mittalCompositionalAttentionDisentangling2022}
Sarthak Mittal, Sharath~Chandra Raparthy, Irina Rish, Yoshua Bengio, and
  Guillaume Lajoie.
\newblock Compositional attention: {{Disentangling}} search and retrieval.
\newblock In {\em International Conference on Learning Representations}, 2022.

\bibitem{mnihAsynchronousMethodsDeep2016}
Volodymyr Mnih, Adri{\`a} Puigdom{\`e}nech~Badia, Mehdi Mirza, Alex Graves,
  Timothy~P. Lillicrap, Tim Harley, David Silver, and Koray Kavukcuoglu.
\newblock Asynchronous {{Methods}} for {{Deep Reinforcement Learning}}.
\newblock {\em arXiv e-prints}, page arXiv:1602.01783, February 2016.

\bibitem{modrakDevelopmentModularityMeasure2021}
Vladimir Modrak and Zuzana Soltysova.
\newblock Development of the {{Modularity Measure}} for {{Assembly Process
  Structures}}.
\newblock {\em Mathematical Problems in Engineering}, 2021:e4900748, December
  2021.

\bibitem{muffLocalModularityMeasure2005}
Stefanie Muff, Francesco Rao, and Amedeo Caflisch.
\newblock Local modularity measure for network clusterizations.
\newblock {\em Physical Review E}, 72(5):056107, November 2005.

\bibitem{murtyCharacterizingIntrinsicCompositionality2022}
Shikhar Murty, Pratyusha Sharma, Jacob Andreas, and Christopher~D. Manning.
\newblock Characterizing {{Intrinsic Compositionality}} in {{Transformers}}
  with {{Tree Projections}}, November 2022.

\bibitem{newmanModularityCommunityStructure2006}
M.~E.~J. Newman.
\newblock Modularity and community structure in networks.
\newblock {\em Proceedings of the National Academy of Sciences},
  103(23):8577--8582, 2006.

\bibitem{opitzEfficientModelAveraging2016}
Michael Opitz, Horst Possegger, and Horst Bischof.
\newblock Efficient model averaging for deep neural networks.
\newblock In {\em Asian Conference on Computer Vision}, pages 205--220.
  {Springer}, 2016.

\bibitem{ostapenkoContinualLearningLocal2021}
Oleksiy Ostapenko, Pau Rodriguez, Massimo Caccia, and Laurent Charlin.
\newblock Continual learning via local module composition.
\newblock {\em Advances in Neural Information Processing Systems},
  34:30298--30312, 2021.

\bibitem{ostapenkoAttentionCompositionalModularity2022}
Oleksiy Ostapenko, Pau Rodriguez, Alexandre Lacoste, and Laurent Charlin.
\newblock Attention for compositional modularity.
\newblock In {\em {{NeurIPS}} '22 Workshop on All Things Attention:
  {{Bridging}} Different Perspectives on Attention}, 2022.

\bibitem{panDecomposingDeepNeural2020}
Rangeet Pan and Hridesh Rajan.
\newblock On {{Decomposing}} a {{Deep Neural Network}} into {{Modules}}.
\newblock In {\em Proceedings of the 28th {{ACM Joint Meeting}} on {{European
  Software Engineering Conference}} and {{Symposium}} on the {{Foundations}} of
  {{Software Engineering}}}, {{ESEC}}/{{FSE}} 2020, pages 889--900, {New York,
  NY, USA}, 2020. {Association for Computing Machinery}.

\bibitem{panDecomposingConvolutionalNeural2021}
Rangeet Pan and Hridesh Rajan.
\newblock Decomposing {{Convolutional Neural Networks}} into {{Reusable}} and
  {{Replaceable Modules}}.
\newblock In {\em Proceedings of {{The}} 44th {{International Conference}} on
  {{Software Engineering}} ({{ICSE}} 2022)}, December 2021.

\bibitem{parascandoloLearningIndependentCausal2018}
Giambattista Parascandolo, Niki Kilbertus, Mateo {Rojas-Carulla}, and Bernhard
  Sch{\"o}lkopf.
\newblock Learning independent causal mechanisms.
\newblock In {\em International Conference on Machine Learning}, pages
  4036--4044. {PMLR}, 2018.

\bibitem{parnasCriteriaBeUsed1972}
D.~L. Parnas.
\newblock On the criteria to be used in decomposing systems into modules.
\newblock {\em Communications of the ACM}, 15(12):1053--1058, December 1972.

\bibitem{paszkePyTorchImperativeStyle2019}
Adam Paszke, Sam Gross, Francisco Massa, Adam Lerer, James Bradbury, Gregory
  Chanan, Trevor Killeen, Zeming Lin, Natalia Gimelshein, Luca Antiga, Alban
  Desmaison, Andreas Kopf, Edward Yang, Zachary DeVito, Martin Raison, Alykhan
  Tejani, Sasank Chilamkurthy, Benoit Steiner, Lu~Fang, Junjie Bai, and Soumith
  Chintala.
\newblock {{PyTorch}}: {{An Imperative Style}}, {{High-Performance Deep
  Learning Library}}.
\newblock In H.~Wallach, H.~Larochelle, A.~Beygelzimer,
  F.~d'{\aftergroup\ignorespaces} {Alch{\'e}-Buc}, E.~Fox, and R.~Garnett,
  editors, {\em Advances in {{Neural Information Processing Systems}} 32},
  pages 8024--8035. {Curran Associates, Inc.}, 2019.

\bibitem{pathakLearningControlSelfassembling2019}
Deepak Pathak, Christopher Lu, Trevor Darrell, Phillip Isola, and Alexei~A
  Efros.
\newblock Learning to control self-assembling morphologies: A study of
  generalization via modularity.
\newblock {\em Advances in Neural Information Processing Systems}, 32, 2019.

\bibitem{pereira-lealOriginsEvolutionFunctional2006}
Jose~B {Pereira-Leal}, Emmanuel~D Levy, and Sarah~A Teichmann.
\newblock The origins and evolution of functional modules: Lessons from protein
  complexes.
\newblock {\em Philosophical Transactions of the Royal Society B: Biological
  Sciences}, 361(1467):507--517, 2006.

\bibitem{petersElementsCausalInference2017}
Jonas Peters, Dominik Janzing, and Bernhard Sch{\"o}lkopf.
\newblock {\em Elements of {{Causal Inference}}: {{Foundations}} and {{Learning
  Algorithms}}}.
\newblock Adaptive {{Computation}} and {{Machine Learning}} Series. {MIT
  Press}, {Cambridge, MA, USA}, November 2017.

\bibitem{poisotPosterioriMeasureNetwork2013}
Timoth{\'e}e Poisot.
\newblock An a posteriori measure of network modularity.
\newblock {\em F1000Research}, 2:130, December 2013.

\bibitem{pontiInductiveBiasModular2021}
Edoardo Ponti.
\newblock {\em Inductive Bias and Modular Design for Sample-Efficient Neural
  Language Learning}.
\newblock PhD thesis, University of Cambridge, 2021.

\bibitem{pontiCombiningModularSkills2022}
Edoardo~M. Ponti, Alessandro Sordoni, Yoshua Bengio, and Siva Reddy.
\newblock Combining {{Modular Skills}} in {{Multitask Learning}}, March 2022.

\bibitem{purushwalkamTaskdrivenModularNetworks2019}
Senthil Purushwalkam, Maximilian Nickel, Abhinav Gupta, and Marc'Aurelio
  Ranzato.
\newblock Task-driven modular networks for zero-shot compositional learning.
\newblock In {\em Proceedings of the {{IEEE}}/{{CVF}} International Conference
  on Computer Vision}, pages 3593--3602, 2019.

\bibitem{pylyshynVisionContinuousCognition1999}
Zenon Pylyshyn.
\newblock Is vision continuous with cognition?: {{The}} case for cognitive
  impenetrability of visual perception.
\newblock {\em Behavioral and Brain Sciences}, 22(3):341--365, June 1999.

\bibitem{qiaoNovelModularRBF2020}
Jun-Fei Qiao, Xi~Meng, Wen-Jing Li, and Bogdan~M. Wilamowski.
\newblock A novel modular {{RBF}} neural network based on a brain-like
  partition method.
\newblock {\em Neural Computing and Applications}, 32(3):899--911, February
  2020.

\bibitem{rahamanDynamicInferenceNeural2021}
Nasim Rahaman, Muhammad~Waleed Gondal, Shruti Joshi, Peter Gehler, Yoshua
  Bengio, Francesco Locatello, and Bernhard Sch{\"o}lkopf.
\newblock Dynamic inference with neural interpreters.
\newblock In M.~Ranzato, A.~Beygelzimer, Y.~Dauphin, P.S. Liang, and J.~Wortman
  Vaughan, editors, {\em Advances in Neural Information Processing Systems},
  volume~34, pages 10985--10998. {Curran Associates, Inc.}, 2021.

\bibitem{ramachandranDiversityDepthPerexample2019}
Prajit Ramachandran and Quoc~V. Le.
\newblock Diversity and depth in per-example routing models.
\newblock In {\em International Conference on Learning Representations}, 2019.

\bibitem{ranganathanStudyFindFacts2021}
G~Ranganathan et~al.
\newblock A study to find facts behind preprocessing on deep learning
  algorithms.
\newblock {\em Journal of Innovative Image Processing (JIIP)}, 3(01):66--74,
  2021.

\bibitem{raviOptimizationModelFewshot2017}
Sachin Ravi and Hugo Larochelle.
\newblock Optimization as a model for few-shot learning.
\newblock In {\em International Conference on Learning Representations}, 2017.

\bibitem{reisingerEvolvingReusableNeural2004}
J.~Reisinger, K.~Stanley, and R.~Miikkulainen.
\newblock Evolving {{Reusable Neural Modules}}.
\newblock In {\em {{GECCO}}}, 2004.

\bibitem{renSurveyDeepActive2021}
Pengzhen Ren, Yun Xiao, Xiaojun Chang, Po-Yao Huang, Zhihui Li, Brij~B Gupta,
  Xiaojiang Chen, and Xin Wang.
\newblock A survey of deep active learning.
\newblock {\em ACM computing surveys (CSUR)}, 54(9):1--40, 2021.

\bibitem{ridgewayLearningDeepDisentangled2018}
Karl Ridgeway and Michael~C Mozer.
\newblock Learning {{Deep Disentangled Embeddings With}} the {{F-Statistic
  Loss}}.
\newblock In {\em Advances in {{Neural Information Processing Systems}}},
  volume~31. {Curran Associates, Inc.}, 2018.

\bibitem{robbinsModularityMind2017}
Philip Robbins.
\newblock Modularity of {{Mind}}.
\newblock In Edward~N. Zalta, editor, {\em The {{Stanford Encyclopedia}} of
  {{Philosophy}}}. {Metaphysics Research Lab, Stanford University}, winter 2017
  edition, 2017.

\bibitem{roseCourseGroupTheory1994}
John~S Rose.
\newblock {\em A Course on Group Theory}.
\newblock {Courier Corporation}, 1994.

\bibitem{rosenbaumRoutingNetworksChallenges2019}
Clemens Rosenbaum, Ignacio Cases, Matthew Riemer, and Tim Klinger.
\newblock Routing {{Networks}} and the {{Challenges}} of {{Modular}} and
  {{Compositional Computation}}, April 2019.

\bibitem{rosenbaumRoutingNetworksAdaptive2018}
Clemens Rosenbaum, Tim Klinger, and Matthew Riemer.
\newblock Routing {{Networks}}: {{Adaptive Selection}} of {{Non-Linear
  Functions}} for {{Multi-Task Learning}}.
\newblock In {\em International {{Conference}} on {{Learning
  Representations}}}, 2018.

\bibitem{ruderOverviewGradientDescent2016}
Sebastian Ruder.
\newblock An overview of gradient descent optimization algorithms.
\newblock {\em arXiv preprint arXiv:1609.04747}, 2016.

\bibitem{rumelhartLearningInternalRepresentations1985}
David~E Rumelhart, Geoffrey~E Hinton, and Ronald~J Williams.
\newblock Learning internal representations by error propagation.
\newblock Technical report, {California Univ San Diego La Jolla Inst for
  Cognitive Science}, 1985.

\bibitem{russakovskyImageNetLargeScale2015}
Olga Russakovsky, Jia Deng, Hao Su, Jonathan Krause, Sanjeev Satheesh, Sean Ma,
  Zhiheng Huang, Andrej Karpathy, Aditya Khosla, Michael Bernstein, et~al.
\newblock {{ImageNet Large Scale Visual Recognition Challenge}}.
\newblock In {\em International Journal of Computer Vision}, volume 115, pages
  211--252. {Springer}, 2015.

\bibitem{rusuProgressiveNeuralNetworks2016}
Andrei~A. Rusu, Neil~C. Rabinowitz, Guillaume Desjardins, Hubert Soyer, James
  Kirkpatrick, Koray Kavukcuoglu, Razvan Pascanu, and Raia Hadsell.
\newblock Progressive {{Neural Networks}}.
\newblock {\em arXiv:1606.04671 [cs]}, September 2016.

\bibitem{salha-galvanModularityAwareGraphAutoencoders2022}
Guillaume {Salha-Galvan}, Johannes~F. Lutzeyer, George Dasoulas, Romain
  Hennequin, and Michalis Vazirgiannis.
\newblock Modularity-{{Aware Graph Autoencoders}} for {{Joint Community
  Detection}} and {{Link Prediction}}, June 2022.

\bibitem{schenkelRecognitionbasedSegmentationOnline1992}
M.~Schenkel, H.~Weissman, I.~Guyon, C.~Nohl, and D.~Henderson.
\newblock Recognition-based segmentation of on-line hand-printed words.
\newblock In S.~Hanson, J.~Cowan, and C.~Giles, editors, {\em Advances in
  Neural Information Processing Systems}, volume~5. {Morgan-Kaufmann}, 1992.

\bibitem{schillingGeneralModularSystems2000}
Melissa Schilling.
\newblock Toward a {{General Modular Systems Theory}} and {{Its Application}}
  to {{Interfirm Product Modularity}}.
\newblock {\em Academy of Management Review}, 25, April 2000.

\bibitem{schmidhuberCompositionalLearningDynamic1990}
J{\"u}rgen Schmidhuber.
\newblock Towards compositional learning in dynamic networks.
\newblock 1990.

\bibitem{schmidtModularityConceptNew2001}
Albrecht Schmidt and Zuhair Bandar.
\newblock Modularity - {{A Concept For New Neural Network Architectures}}.
\newblock November 2001.

\bibitem{shaoModularityMeasuresConcepts2020}
Yue Shao and Victor~M. Zavala.
\newblock Modularity measures: {{Concepts}}, computation, and applications to
  manufacturing systems.
\newblock {\em AIChE Journal}, 66(6):e16965, 2020.

\bibitem{shazeerOutrageouslyLargeNeural2017}
Noam Shazeer, Azalia Mirhoseini, Krzysztof Maziarz, Andy Davis, Quoc~V. Le,
  Geoffrey~E. Hinton, and Jeff Dean.
\newblock Outrageously {{Large Neural Networks}}: {{The Sparsely-Gated
  Mixture-of-Experts Layer}}.
\newblock In {\em 5th {{International Conference}} on {{Learning
  Representations}}, {{ICLR}} 2017, {{Toulon}}, {{France}}, {{April}} 24-26,
  2017, {{Conference Track Proceedings}}}. {OpenReview.net}, 2017.

\bibitem{shiScriptIdentificationWild2016}
Baoguang Shi, Xiang Bai, and Cong Yao.
\newblock Script identification in the wild via discriminative convolutional
  neural network.
\newblock {\em Pattern Recognition}, 52:448--458, April 2016.

\bibitem{shinContinualLearningDeep2017}
Hanul Shin, Jung~Kwon Lee, Jaehong Kim, and Jiwon Kim.
\newblock Continual {{Learning}} with {{Deep Generative Replay}}.
\newblock In I.~Guyon, U.~V. Luxburg, S.~Bengio, H.~Wallach, R.~Fergus,
  S.~Vishwanathan, and R.~Garnett, editors, {\em Advances in {{Neural
  Information Processing Systems}}}, volume~30. {Curran Associates, Inc.},
  2017.

\bibitem{shiokawaFastAlgorithmModularitybased2013}
Hiroaki Shiokawa, Yasuhiro Fujiwara, and Makoto Onizuka.
\newblock Fast algorithm for modularity-based graph clustering.
\newblock In {\em Proceedings of the {{AAAI Conference}} on {{Artificial
  Intelligence}}}, volume~27, pages 1170--1176, 2013.

\bibitem{sifreRigidmotionScatteringImage2014}
L~Sifre.
\newblock {\em Rigid-Motion Scattering for Image Classification [{{PhD}}
  Thesis]}.
\newblock PhD thesis, 2014.

\bibitem{silverMasteringGameGo2016}
David Silver, Aja Huang, Chris~J. Maddison, Arthur Guez, Laurent Sifre, George
  {van den Driessche}, Julian Schrittwieser, Ioannis Antonoglou, Veda
  Panneershelvam, Marc Lanctot, Sander Dieleman, Dominik Grewe, John Nham, Nal
  Kalchbrenner, Ilya Sutskever, Timothy Lillicrap, Madeleine Leach, Koray
  Kavukcuoglu, Thore Graepel, and Demis Hassabis.
\newblock Mastering the game of {{Go}} with deep neural networks and tree
  search.
\newblock {\em Nature}, 529(7587):484--489, January 2016.

\bibitem{silverMasteringGameGo2017}
David Silver, Julian Schrittwieser, Karen Simonyan, Ioannis Antonoglou, Aja
  Huang, Arthur Guez, Thomas Hubert, Lucas Baker, Matthew Lai, Adrian Bolton,
  Yutian Chen, Timothy Lillicrap, Fan Hui, Laurent Sifre, George {van den
  Driessche}, Thore Graepel, and Demis Hassabis.
\newblock Mastering the game of {{Go}} without human knowledge.
\newblock {\em Nature}, 550(7676):354--359, October 2017.

\bibitem{simardBestPracticesConvolutional2003}
P.~Y. Simard, D.~Steinkraus, and J.~C. Platt.
\newblock Best practices for convolutional neural networks applied to visual
  document analysis.
\newblock In {\em Seventh {{International Conference}} on {{Document Analysis}}
  and {{Recognition}}, 2003. {{Proceedings}}.}, pages 958--963, August 2003.

\bibitem{simonArchitectureComplexity1962}
Herbert~A. Simon.
\newblock The {{Architecture}} of {{Complexity}}.
\newblock {\em Proceedings of the American Philosophical Society},
  106(6):467--482, 1962.

\bibitem{simonAggregationVariablesDynamic1961}
Herbert~A. Simon and Albert Ando.
\newblock Aggregation of variables in dynamic systems.
\newblock {\em Econometrica}, 29(2):111--138, 1961.

\bibitem{simpkinsComposableModularReinforcement2019}
Christopher Simpkins and Charles Isbell.
\newblock Composable modular reinforcement learning.
\newblock In {\em Proceedings of the {{AAAI}} Conference on Artificial
  Intelligence}, volume~33, pages 4975--4982, 2019.

\bibitem{sinitsinEditableNeuralNetworks2019}
Anton Sinitsin, Vsevolod Plokhotnyuk, Dmitry Pyrkin, Sergei Popov, and Artem
  Babenko.
\newblock Editable {{Neural Networks}}.
\newblock In {\em International {{Conference}} on {{Learning
  Representations}}}, 2019.

\bibitem{smithDonDecayLearning2018}
Samuel~L Smith, Pieter-Jan Kindermans, Chris Ying, and Quoc~V Le.
\newblock Don't decay the learning rate, increase the batch size.
\newblock In {\em International Conference on Learning Representations}, 2018.

\bibitem{smithUsingDeepSpeedMegatron2022}
Shaden Smith, Mostofa Patwary, Brandon Norick, Patrick LeGresley, Samyam
  Rajbhandari, Jared Casper, Zhun Liu, Shrimai Prabhumoye, George Zerveas,
  Vijay Korthikanti, Elton Zhang, Rewon Child, Reza~Yazdani Aminabadi, Julie
  Bernauer, Xia Song, Mohammad Shoeybi, Yuxiong He, Michael Houston, Saurabh
  Tiwary, and Bryan Catanzaro.
\newblock Using {{DeepSpeed}} and {{Megatron}} to {{Train Megatron-Turing NLG
  530B}}, {{A Large-Scale Generative Language Model}}, February 2022.

\bibitem{snellPrototypicalNetworksFewshot2017}
Jake Snell, Kevin Swersky, and Richard~S. Zemel.
\newblock Prototypical {{Networks}} for {{Few-shot Learning}}.
\newblock {\em arXiv:1703.05175 [cs, stat]}, June 2017.

\bibitem{srivastavaDropoutSimpleWay2014}
Nitish Srivastava, Geoffrey Hinton, Alex Krizhevsky, Ilya Sutskever, and Ruslan
  Salakhutdinov.
\newblock Dropout: {{A}} simple way to prevent neural networks from
  overfitting.
\newblock {\em Journal of Machine Learning Research}, 15(56):1929--1958, 2014.

\bibitem{sunRevisitingUnreasonableEffectiveness2017}
Chen Sun, Abhinav Shrivastava, Saurabh Singh, and Abhinav Gupta.
\newblock Revisiting unreasonable effectiveness of data in deep learning era.
\newblock In {\em Proceedings of the {{IEEE}} International Conference on
  Computer Vision}, pages 843--852, 2017.

\bibitem{sunTaskSwitchingNetwork2021}
Guolei Sun, Thomas Probst, Danda Pani~Paudel, Nikola Popovic, Menelaos Kanakis,
  Jagruti Patel, Dengxin Dai, and Luc Van~Gool.
\newblock Task {{Switching Network}} for {{Multi-task Learning}}.
\newblock In {\em 2021 {{IEEE}}/{{CVF International Conference}} on {{Computer
  Vision}} ({{ICCV}})}, pages 8271--8280, {Montreal, QC, Canada}, October 2021.
  {IEEE}.

\bibitem{sunOmniPrintConfigurablePrinted2021}
Haozhe Sun, Wei-Wei Tu, and Isabelle~M. Guyon.
\newblock {{OmniPrint}}: {{A Configurable Printed Character Synthesizer}}.
\newblock In {\em Thirty-Fifth {{Conference}} on {{Neural Information
  Processing Systems Datasets}} and {{Benchmarks Track}} ({{Round}} 1)}, 2021.

\bibitem{suttonReinforcementLearningIntroduction2018}
Richard~S. Sutton and Andrew~G. Barto.
\newblock {\em Reinforcement {{Learning}}: {{An Introduction}}}.
\newblock {The MIT Press}, second edition, 2018.

\bibitem{szegedyInceptionv4InceptionresnetImpact2017}
Christian Szegedy, Sergey Ioffe, Vincent Vanhoucke, and Alexander~A Alemi.
\newblock Inception-v4, inception-resnet and the impact of residual connections
  on learning.
\newblock In {\em Thirty-First {{AAAI}} Conference on Artificial Intelligence},
  2017.

\bibitem{szegedyRethinkingInceptionArchitecture2016}
Christian Szegedy, Vincent Vanhoucke, Sergey Ioffe, Jon Shlens, and Zbigniew
  Wojna.
\newblock Rethinking the inception architecture for computer vision.
\newblock In {\em Proceedings of the {{IEEE}} Conference on Computer Vision and
  Pattern Recognition}, pages 2818--2826, 2016.

\bibitem{tarvainenMeanTeachersAre2018}
Antti Tarvainen and Harri Valpola.
\newblock Mean teachers are better role models: {{Weight-averaged}} consistency
  targets improve semi-supervised deep learning results.
\newblock {\em arXiv:1703.01780 [cs, stat]}, April 2018.

\bibitem{teerapittayanonBranchyNetFastInference2016}
Surat Teerapittayanon, Bradley McDanel, and H.T. Kung.
\newblock {{BranchyNet}}: {{Fast}} inference via early exiting from deep neural
  networks.
\newblock In {\em 2016 23rd {{International Conference}} on {{Pattern
  Recognition}} ({{ICPR}})}, pages 2464--2469, 2016.

\bibitem{terekhovKnowledgeTransferDeep2015}
Alexander Terekhov, Guglielmo Montone, and J.~O'Regan.
\newblock Knowledge {{Transfer}} in {{Deep Block-Modular Neural Networks}}.
\newblock In {\em Biomimetic and {{Biohybrid Systems}}}, pages 268--279.
  {Springer}, July 2015.

\bibitem{tishbyDeepLearningInformation2015}
Naftali Tishby and Noga Zaslavsky.
\newblock Deep learning and the information bottleneck principle.
\newblock In {\em 2015 {{IEEE Information Theory Workshop}} ({{ITW}})}, pages
  1--5, 2015.

\bibitem{triantafillouMetadatasetDatasetDatasets2019}
Eleni Triantafillou, Tyler Zhu, Vincent Dumoulin, Pascal Lamblin, Utku Evci,
  Kelvin Xu, Ross Goroshin, Carles Gelada, Kevin Swersky, Pierre-Antoine
  Manzagol, et~al.
\newblock Meta-dataset: {{A}} dataset of datasets for learning to learn from
  few examples.
\newblock In {\em International Conference on Learning Representations}, 2019.

\bibitem{ullahMetaAlbumMultidomainMetaDataset2022}
Ihsan Ullah, Dustin Carrion, Sergio Escalera, Isabelle~M. Guyon, Mike Huisman,
  Felix Mohr, Jan~N. van Rijn, Haozhe Sun, Joaquin Vanschoren, and Phan~Anh Vu.
\newblock Meta-{{Album}}: {{Multi-domain Meta-Dataset}} for {{Few-Shot Image
  Classification}}, 2022.

\bibitem{vankovTrainingNeuralNetworks2020}
Ivan~I Vankov and Jeffrey~S Bowers.
\newblock Training neural networks to encode symbols enables combinatorial
  generalization.
\newblock {\em Philosophical Transactions of the Royal Society B},
  375(1791):20190309, 2020.

\bibitem{vaswaniAttentionAllYou2017}
Ashish Vaswani, Noam Shazeer, Niki Parmar, Jakob Uszkoreit, Llion Jones,
  Aidan~N Gomez, {\L}ukasz Kaiser, and Illia Polosukhin.
\newblock Attention is all you need.
\newblock In I.~Guyon, U.~Von Luxburg, S.~Bengio, H.~Wallach, R.~Fergus,
  S.~Vishwanathan, and R.~Garnett, editors, {\em Advances in Neural Information
  Processing Systems}, volume~30. {Curran Associates, Inc.}, 2017.

\bibitem{veniatEfficientContinualLearning2021}
Tom Veniat, Ludovic Denoyer, and Marc'Aurelio Ranzato.
\newblock Efficient {{Continual Learning}} with {{Modular Networks}} and
  {{Task-Driven Priors}}.
\newblock In {\em 9th {{International Conference}} on {{Learning
  Representations}}, {{ICLR}} 2021}, 2021.

\bibitem{luxburgClusteringScienceArt2012}
Ulrike von Luxburg, Robert~C. Williamson, and Isabelle Guyon.
\newblock Clustering: {{Science}} or {{Art}}?
\newblock In {\em Proceedings of {{ICML Workshop}} on {{Unsupervised}} and
  {{Transfer Learning}}}, pages 65--79. {JMLR Workshop and Conference
  Proceedings}, June 2012.

\bibitem{wagnerPerspectiveComplexAdaptations1996}
Gunter~P. Wagner and Lee Altenberg.
\newblock Perspective: {{Complex Adaptations}} and the {{Evolution}} of
  {{Evolvability}}.
\newblock {\em Evolution}, 50(3):967--976, 1996.

\bibitem{wangBridgingMultitaskLearning2021}
Haoxiang Wang, Han Zhao, and Bo~Li.
\newblock Bridging multi-task learning and meta-learning: {{Towards}} efficient
  training and effective adaptation.
\newblock In {\em International Conference on Machine Learning}, pages
  10991--11002. {PMLR}, 2021.

\bibitem{wangCombinatorialPerspectiveTransfer2020}
Jianan Wang, Eren Sezener, David Budden, Marcus Hutter, and Joel Veness.
\newblock A {{Combinatorial Perspective}} on {{Transfer Learning}}.
\newblock In H.~Larochelle, M.~Ranzato, R.~Hadsell, M.~F. Balcan, and H.~Lin,
  editors, {\em Advances in {{Neural Information Processing Systems}}},
  volume~33, pages 918--929. {Curran Associates, Inc.}, 2020.

\bibitem{wangRoleGlobalLabels2021}
Ruohan Wang, Massimiliano Pontil, and Carlo Ciliberto.
\newblock The role of global labels in few-shot classification and how to infer
  them.
\newblock In {\em Advances in {{Neural Information Processing Systems}}},
  volume~34, pages 27160--27170, 2021.

\bibitem{watanabeModularRepresentationLayered2018}
Chihiro Watanabe, Kaoru Hiramatsu, and Kunio Kashino.
\newblock Modular representation of layered neural networks.
\newblock {\em Neural Networks}, 97:62--73, 2018.

\bibitem{weilerGeneralEquivariantSteerable2021}
Maurice Weiler and Gabriele Cesa.
\newblock General \${{E}}(2)\$-{{Equivariant Steerable CNNs}}.
\newblock {\em arXiv:1911.08251 [cs, eess]}, April 2021.

\bibitem{weilerLearningSteerableFilters2018}
Maurice Weiler, Fred~A. Hamprecht, and Martin Storath.
\newblock Learning {{Steerable Filters}} for {{Rotation Equivariant CNNs}}.
\newblock {\em arXiv:1711.07289 [cs]}, March 2018.

\bibitem{worrallHarmonicNetworksDeep2017}
Daniel~E. Worrall, Stephan~J. Garbin, Daniyar Turmukhambetov, and Gabriel~J.
  Brostow.
\newblock Harmonic {{Networks}}: {{Deep Translation}} and {{Rotation
  Equivariance}}.
\newblock In {\em Proceedings of the {{IEEE Conference}} on {{Computer Vision}}
  and {{Pattern Recognition}}}, pages 5028--5037, 2017.

\bibitem{wuLearningImplicitSemantic2021}
Likang Wu, Zhi Li, Hongke Zhao, Qi~Liu, Jun Wang, Mengdi Zhang, and Enhong
  Chen.
\newblock Learning the implicit semantic representation on graph-structured
  data.
\newblock In {\em International Conference on Database Systems for Advanced
  Applications}, pages 3--19. {Springer}, 2021.

\bibitem{wuScalableTrustregionMethod2017}
Yuhuai Wu, Elman Mansimov, Shun Liao, Roger Grosse, and Jimmy Ba.
\newblock Scalable trust-region method for deep reinforcement learning using
  {{Kronecker-factored}} approximation.
\newblock {\em arXiv e-prints}, page arXiv:1708.05144, August 2017.

\bibitem{xieAggregatedResidualTransformations2017}
Saining Xie, Ross Girshick, Piotr Dollar, Zhuowen Tu, and Kaiming He.
\newblock Aggregated {{Residual Transformations}} for {{Deep Neural Networks}}.
\newblock In {\em Proceedings of the {{IEEE Conference}} on {{Computer Vision}}
  and {{Pattern Recognition}} ({{CVPR}})}, July 2017.

\bibitem{xieExploringRandomlyWired2019}
Saining Xie, Alexander Kirillov, Ross Girshick, and Kaiming He.
\newblock Exploring {{Randomly Wired Neural Networks}} for {{Image
  Recognition}}.
\newblock In {\em Proceedings of the {{IEEE}}/{{CVF International Conference}}
  on {{Computer Vision}} ({{ICCV}})}, October 2019.

\bibitem{xiongConditionalConvolutionalNeural2015}
Chao Xiong, Xiaowei Zhao, Danhang Tang, Karlekar Jayashree, Shuicheng Yan, and
  Tae-Kyun Kim.
\newblock Conditional convolutional neural network for modality-aware face
  recognition.
\newblock In {\em Proceedings of the {{IEEE International Conference}} on
  {{Computer Vision}}}, pages 3667--3675, 2015.

\bibitem{yalnizBillionscaleSemisupervisedLearning2019a}
I.~Zeki Yalniz, Herv{\'e} J{\'e}gou, Kan Chen, Manohar Paluri, and Dhruv
  Mahajan.
\newblock Billion-scale semi-supervised learning for image classification.
\newblock {\em CoRR}, abs/1905.00546, 2019.

\bibitem{yangLSTMGRUNeural2020}
Shudong Yang, Xueying Yu, and Ying Zhou.
\newblock {{LSTM}} and {{GRU}} neural network performance comparison study:
  {{Taking}} yelp review dataset as an example.
\newblock In {\em 2020 International Workshop on Electronic Communication and
  Artificial Intelligence ({{IWECAI}})}, pages 98--101, 2020.

\bibitem{yaoHierarchicalMixtureClassification2009}
Bangpeng Yao, Dirk Walther, Diane Beck, and Li~{Fei-fei}.
\newblock Hierarchical mixture of classification experts uncovers interactions
  between brain regions.
\newblock In Y.~Bengio, D.~Schuurmans, J.~Lafferty, C.~Williams, and
  A.~Culotta, editors, {\em Advances in Neural Information Processing Systems},
  volume~22. {Curran Associates, Inc.}, 2009.

\bibitem{yingNasbench101ReproducibleNeural2019}
Chris Ying, Aaron Klein, Eric Christiansen, Esteban Real, Kevin Murphy, and
  Frank Hutter.
\newblock Nas-bench-101: {{Towards}} reproducible neural architecture search.
\newblock In {\em International Conference on Machine Learning}, pages
  7105--7114. {PMLR}, 2019.

\bibitem{yosinskiHowTransferableAre2014}
Jason Yosinski, Jeff Clune, Yoshua Bengio, and Hod Lipson.
\newblock How transferable are features in deep neural networks?
\newblock {\em arXiv:1411.1792 [cs]}, November 2014.

\bibitem{yuSlimmableNeuralNetworks2019}
Jiahui Yu, Linjie Yang, Ning Xu, Jianchao Yang, and Thomas Huang.
\newblock Slimmable {{Neural Networks}}.
\newblock In {\em International {{Conference}} on {{Learning
  Representations}}}, 2019.

\bibitem{yuMAttNetModularAttention2018}
Licheng Yu, Zhe Lin, Xiaohui Shen, Jimei Yang, Xin Lu, Mohit Bansal, and
  Tamara~L Berg.
\newblock {{MAttNet}}: {{Modular}} attention network for referring expression
  comprehension.
\newblock In {\em Proceedings of the {{IEEE}} Conference on Computer Vision and
  Pattern Recognition}, pages 1307--1315, 2018.

\bibitem{yuGradientSurgeryMultiTask2020}
Tianhe Yu, Saurabh Kumar, Abhishek Gupta, Sergey Levine, Karol Hausman, and
  Chelsea Finn.
\newblock Gradient {{Surgery}} for {{Multi-Task Learning}}.
\newblock In {\em Advances in {{Neural Information Processing Systems}}},
  volume~33, pages 5824--5836. {Curran Associates, Inc.}, 2020.

\bibitem{zagoruykoWideResidualNetworks2016}
Sergey Zagoruyko and Nikos Komodakis.
\newblock Wide residual networks.
\newblock {\em arXiv preprint arXiv:1605.07146}, 2016.

\bibitem{zaidiMeasuringDisentanglementReview2021}
Julian Zaidi, Jonathan Boilard, Ghyslain Gagnon, and Marc-Andr{\'e} Carbonneau.
\newblock Measuring {{Disentanglement}}: {{A Review}} of {{Metrics}}.
\newblock {\em arXiv:2012.09276 [cs]}, January 2021.

\bibitem{zhangNetworkTransplanting2018}
Quanshi Zhang, Yu~Yang, Qian Yu, and Ying~Nian Wu.
\newblock Network {{Transplanting}}.
\newblock {\em arXiv:1804.10272 [cs, stat]}, December 2018.

\bibitem{zhangSurveyMultitaskLearning2021}
Yu~Zhang and Qiang Yang.
\newblock A survey on multi-task learning.
\newblock {\em IEEE Transactions on Knowledge and Data Engineering}, 2021.

\bibitem{zhouMetaLearningSymmetriesReparameterization2020}
Allan Zhou, Tom Knowles, and Chelsea Finn.
\newblock Meta-{{Learning Symmetries}} by {{Reparameterization}}.
\newblock {\em arXiv:2007.02933 [cs, stat]}, October 2020.

\bibitem{zhouDiverseEnsembleEvolution2018}
Tianyi Zhou, Shengjie Wang, and Jeff~A Bilmes.
\newblock Diverse ensemble evolution: {{Curriculum}} data-model marriage.
\newblock In S.~Bengio, H.~Wallach, H.~Larochelle, K.~Grauman,
  N.~{Cesa-Bianchi}, and R.~Garnett, editors, {\em Advances in Neural
  Information Processing Systems}, volume~31. {Curran Associates, Inc.}, 2018.

\bibitem{zhouEnsembleMethodsFoundations2012}
Zhi-Hua Zhou.
\newblock {\em Ensemble Methods: Foundations and Algorithms}.
\newblock {CRC press}, 2012.

\bibitem{zhuUnpairedImagetoImageTranslation2017}
Jun-Yan Zhu, Taesung Park, Phillip Isola, and Alexei~A Efros.
\newblock Unpaired {{Image-to-Image Translation}} using {{Cycle-Consistent
  Adversarial Networks}}.
\newblock In {\em Computer {{Vision}} ({{ICCV}}), 2017 {{IEEE International
  Conference}} On}, 2017.

\bibitem{zophLearningTransferableArchitectures2018}
Barret Zoph, Vijay Vasudevan, Jonathon Shlens, and Quoc~V. Le.
\newblock Learning {{Transferable Architectures}} for {{Scalable Image
  Recognition}}.
\newblock {\em arXiv:1707.07012 [cs, stat]}, April 2018.

\end{thebibliography}
}

\end{document}